\documentclass[10pt,twocolumn,letterpaper]{article}

\usepackage{cvpr}              

\usepackage{graphicx}
\usepackage{amsmath}
\usepackage{amssymb}
\usepackage{booktabs}
\usepackage{algorithm}
\usepackage{algorithmic}
\usepackage{epstopdf}
\usepackage{multirow}

\makeatletter
\usepackage[pagebackref,breaklinks,colorlinks]{hyperref}

\usepackage[capitalize]{cleveref}
\crefname{section}{Sec.}{Secs.}
\Crefname{section}{Section}{Sections}
\Crefname{table}{Table}{Tables}
\crefname{table}{Tab.}{Tabs.}

\def\cvprPaperID{7303} 
\def\confName{CVPR}
\def\confYear{2023}

\makeatletter
\def\thanks#1{\protected@xdef\@thanks{\@thanks
        \protect\footnotetext{#1}}}
\makeatother
\begin{document}
\sloppy
\title{DCS-RISR: Dynamic Channel Splitting for Efficient Real-world Image Super-Resolution}
\renewcommand{\thefootnote}{\fnsymbol{footnote}}
\thanks{* Corresponding author (e-mail: shlin@cs.ecnu.edu.cn)}
\author{Junbo Qiao{$^{1}$}, Shaohui Lin{$^{1,*}$}, Yunlun Zhang{$^{2}$}, Wei Li{$^{3}$}, Jie Hu{$^{3}$}, Gaoqi He{$^{1}$}, Changbo Wang{$^{1}$}, Lizhuang Ma{$^{1}$}\\
{$^{1}$}East China Normal University, {$^{2}$}ETH Zürich, {$^{3}$}Huawei Technologies Ltd\\
\tt\small 	\{georgelucien667, yulun100\}@gmail.com, 	\{wei.lee,	 hujie23\}@huawei.com,\\
\tt\small\{shlin, gqhe, cbwang, lzma\}@cs.ecnu.edu.cn
}

\maketitle

\begin{abstract}
Real-world image super-resolution (RISR) has received increased focus for improving the quality of SR images under unknown complex degradation. Existing methods rely on the heavy SR models to enhance low-resolution (LR) images of different degradation levels, which significantly restricts their practical deployments on resource-limited devices.
In this paper, we propose a novel Dynamic Channel Splitting scheme for efficient Real-world Image Super-Resolution, termed DCS-RISR.
Specifically, we first introduce the light degradation prediction network to regress the degradation vector to simulate the real-world degradations, upon which the channel splitting vector is generated as the input for an efficient SR model. Then, a learnable octave convolution block is proposed to adaptively decide the channel splitting scale for low- and high-frequency features at each block, reducing computation overhead and memory cost by offering the large scale to low-frequency features and the small scale to the high ones. To further improve the RISR performance, Non-local regularization is employed to supplement the knowledge of patches from LR and HR subspace with free-computation inference. Extensive experiments demonstrate the effectiveness of DCS-RISR on different benchmark datasets. Our DCS-RISR not only achieves the best trade-off between computation/parameter and PSNR/SSIM metric, and also effectively handles real-world images with different degradation levels.   
\end{abstract}

\begin{figure}[t]
\begin{minipage}{0.5\linewidth}
\vspace{0.3cm}
\centerline{\includegraphics[width=\textwidth]{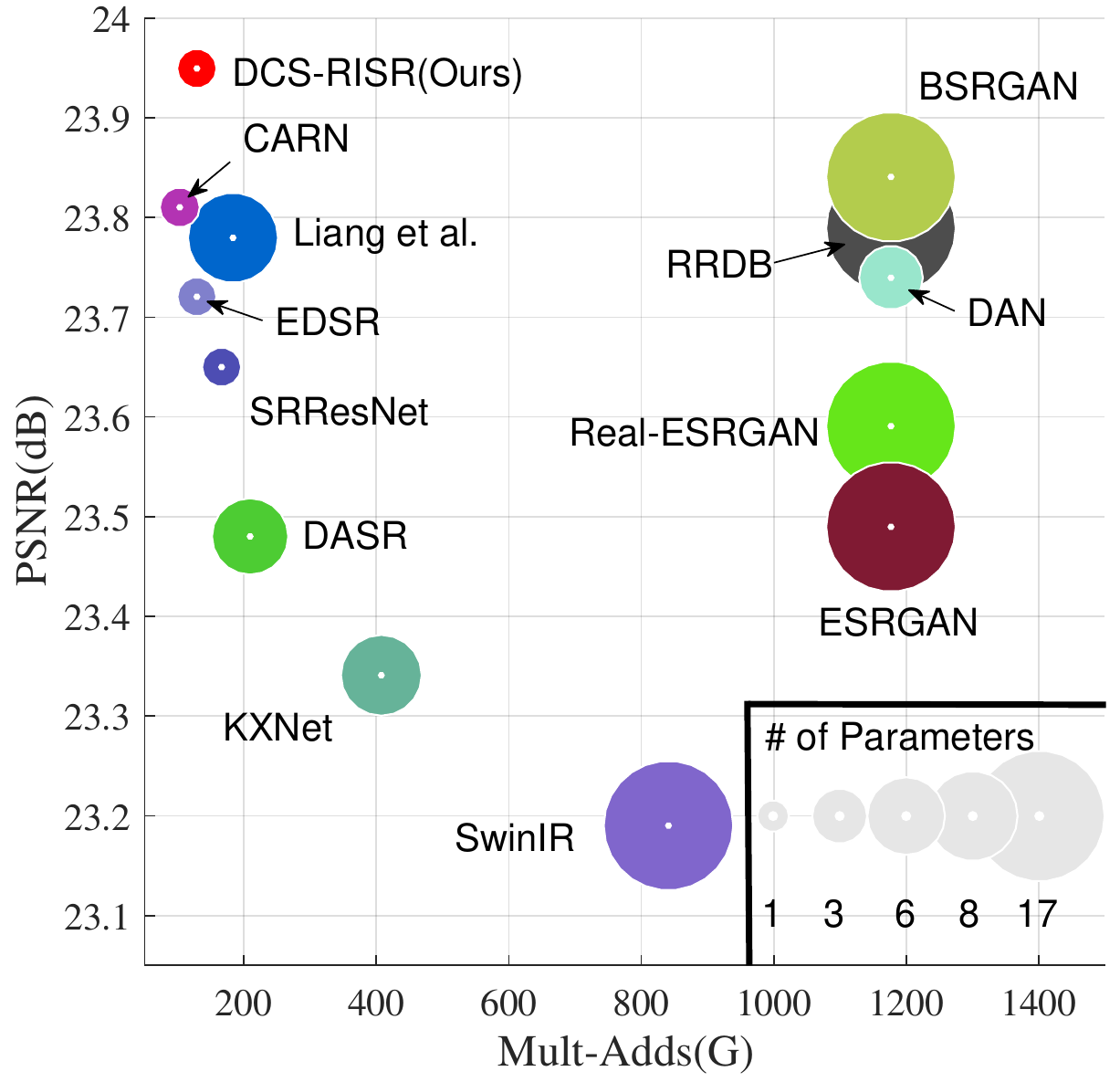}}
\end{minipage}
\begin{minipage}{0.21\linewidth}
\centerline{\includegraphics[width=\textwidth]{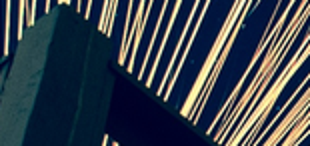}}
\vspace{-0.2cm}
\centerline{\tiny HR 0828.png}
\centerline{\includegraphics[width=\textwidth]{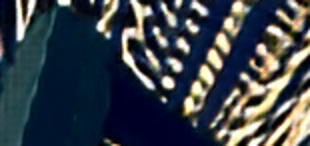}}
\vspace{-0.2cm}
\centerline{\tiny EDSR 0828.png}
\centerline{\includegraphics[width=\textwidth]{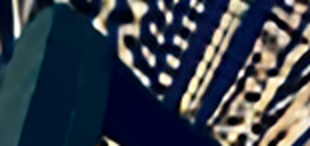}}
\vspace{-0.2cm}
\centerline{\tiny KXNet 0828.png}
\end{minipage}
\begin{minipage}{0.21\linewidth}
\centerline{\includegraphics[width=\textwidth]{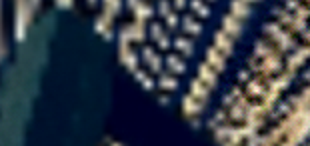}}
\vspace{-0.2cm}
\centerline{\tiny SRResNet 0828.png}
\centerline{\includegraphics[width=\textwidth]{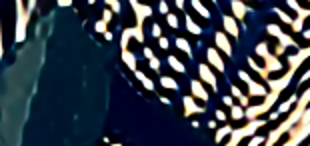}}
\vspace{-0.2cm}
\centerline{\tiny CARN 0828.png}
\centerline{\includegraphics[width=\textwidth]{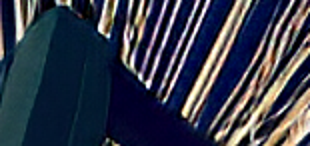}}
\vspace{-0.2cm}
\centerline{\tiny DCS-RISR(Ours)}
\end{minipage}
\centerline{\small (a)\hspace{3.7cm}(b)}
\caption{Motivations of visualization and metric comparison on the DIV2K degradation level 1 for 4x SR. (a) Metric evaluation. (b) Visual comparison.}
\label{fig_mov}
\end{figure}

\section{Introduction}
\label{sec:intro}

Single image super-resolution (SISR)~\cite{glasner2009super,dong2014learning} aims to recover the natural high resolution (HR) image from a single low resolution (LR) image, which has been actively explored due to its high practical applications (\emph{e.g.}, image/video quality enhancement on televisions and medical imaging). 
Convolution neural networks (CNNs)~\cite{wang2018esrgan,lim2017enhanced,zhang2018image,soh2019natural} have achieved remarkable results in the research of SISR due to their powerful feature representation capability. 
However, these methods assume that the degradation from HR to LR images is the bicubic downsampling, which leads to a severe performance drop to super-resolve real-world LR images due to the mismatch degradation.
%

To alleviate the mismatch degradation problem, blind image super-resolution (BISR) \cite{huang2020unfolding,zhang2018learning,zhou2019kernel,gu2019blind,shocher2018zero} has been proposed to restore LR images suffering from unknown and more complex degradations. For example, a simple combination of multiple degradations~\cite{bell2019blind,zhang2018learning} involving blur, noise and downsampling is learned and leveraged into non-blind network to produce a HR output. However, these methods suffer from serious performance drop~\cite{efrat2013accurate}, when the the kernel and noise levels are not accurately estimated. To better estimate the degradation parameters, alternative optimization~\cite{gu2019blind,huang2020unfolding,luo2022learning,fu2022kxnet} between blur kernel estimation and intermediate SR images has been proposed to progressively refine the blur kernel and SR images by mutual learning. Alternatively, contrastive learning~\cite{wang2021unsupervised,zhang2021blind} is introduced to learn degradation-aware representation, which is flexible to adapt to various degradations. They, however, still easily fail in the real-world SR due to the more complex real-world degradations, such as the complicate combination affection of camera blur, sensor noise, JPEG compression, \emph{etc.}

\begin{figure*}[t]
\centering
\includegraphics[width=6.8 in]{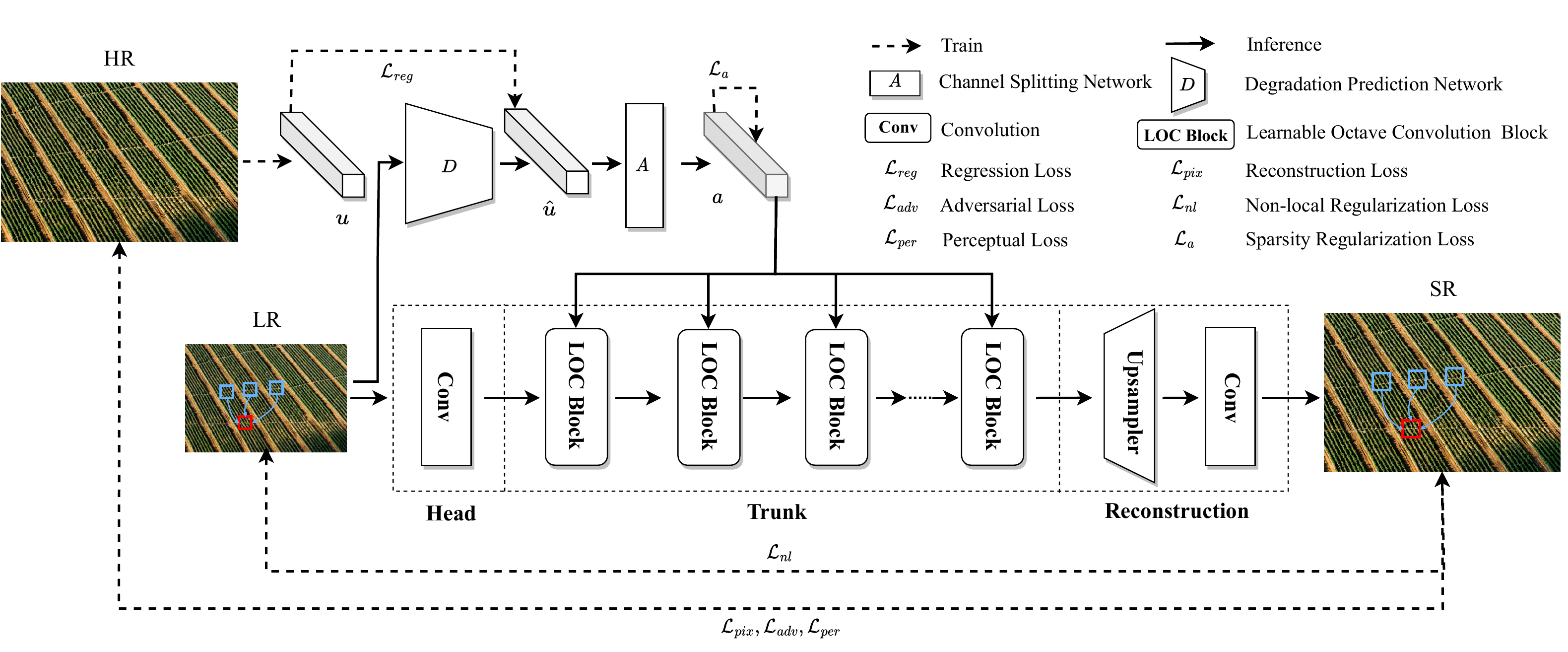}
\vspace{-0.5em}
\caption{Illustration of the proposed DCS-RISR, which contains five key components, degradation prediction network $D$, channel splitting network $A$, Learnable Octave convolution (LOC) block, sparsity regularizer $\mathcal{L}_a$ and non-local regularization $\mathcal{L}_{nl}$.}
\label{fig_2}
\end{figure*}

To be real-world SR, recent RISR tends to explore more complicate combinations of different degradation processes (\emph{e.g.}, random shuffle of degradation orders in BSRGAN~\cite{zhang2021designing} and second-order degradation in Real-ESRGAN~\cite{wang2021real}), to restore real-world LR images by training on the synthesized pairs with more practical degradations.
However, the above RISR methods rely on high-capacity backbone networks (\emph{e.g.}, RRDB~\cite{wang2018esrgan}), multi-stage restoration with heavy computation cost~\cite{gu2019blind,huang2020unfolding}, or significant memory storage~\cite{liang2022efficient}, which restrict their usage on resource-limited devices (shown in Fig.~\ref{fig_mov}(a)). 
Naturally, model compression has become a feasible way to reduce the computation overhead and memory footprint for efficient SR, such as compact SR block design~\cite{ahn2018fast,song2021addersr,nie2021ghostsr}, quantization~\cite{li2020pams,hong2022cadyq} and dynamic patch-based inference~\cite{chen2022arm,wang2022adaptive,kong2021classsr}. 
However, these methods built on the assumption of bicubic degradation, which leads to the failure for the real-world image super-resolution with complex degradation. As shown in Fig.~\ref{fig_mov}(b), CARN~\cite{ahn2018fast} and KXNet~\cite{fu2022kxnet} produce visual artifact and messy details to super-resolve the image from DIV2K with complex level-1\renewcommand{\thefootnote}{\arabic{footnote}}\footnote[1]{It is a joint combination of different degradations well defined in~\cite{liang2022efficient}.} degradation, respectively. This begs our rethinking: \emph{Why not leverage the explicit degradation information into efficient SR block through effective learning strategy for efficient real-world image super-resolution?}


%


To answer the above question, we propose a novel \emph{Dynamic Channel Splitting} scheme for efficient Real-world Image Super-Resolution, termed DCS-RISR, which simultaneously considers the real-world complex degradation and model redundancy. Fig.~\ref{fig_2} depicts the workflow of the proposed approach. Inspired by Real-ESRGAN ~\cite{wang2021real}, we first introduce the multi-level degradation to construct the LR-HR pairs and degradation vector $\mathbf{d}$, which simulates the real-world degradations. Light degradation prediction network $D$ is then used to explicitly regress the degradation vector, upon which channel splitting vector $\mathbf{a}$ is generated through channel splitting network (\emph{e.g.}, MLP) $A$ to adaptively allocate different weights for the channels of efficient SR network.
Specifically, inspired by~\cite{chen2019drop} on the relationship between frequency and redundancy, we design \emph{Learnable Octave Convolution} (LOC) block as our main SR block to dynamically allocate channel splitting scales for high- and low-frequency features, and reduce the spatial size of low-frequency features. As such, we reduce the SR network redundancy by allocating a small amount of computation to high-frequency features if the LR image is smooth with a simple pattern.
%
Furthermore, we propose \emph{non-local regularization} to exploit the relationship of the similar-pattern patches between LR and SR subspace, which remedies the low reconstruction ability on a simple L1-norm reconstruction loss to improve the model performance without additional inference burden. 

We summarize our main contributions as follows: 
\begin{itemize}
    \item We propose a dynamic channel splitting (DCS) to efficiently and effectively super-resolve the real-world LR images. It is able to adaptively adjust the low- and high-frequency proportion of each layer's feature to address the problems of practical degradation and model redundancy.
    \item Learnable octave convolution (LOC) block is compact to add the off-the-shelf SR models for dynamic inference, non-local regularization is employed to supplement the knowledge of patches from LR and SR subspace, which benefits from the performance improvement with free-computation inference.
    \item Extensive experiments demonstrate the superior performance of our approach on various datasets. On DIV2K with the degradation level of 3, the proposed DCS-RISR achieves 23.95dB and 128 GFLOPs outperforming state-of-the-art methods.
\end{itemize}

\section{Related Work}
\label{sec:Rel}

\subsection{Real-world Image Super-Resolution}

CNNs have been widely-used for SISR, such as SRCNN~\cite{dong2014learning}, EDSR~\cite{lim2017enhanced}, SRResNet~\cite{ledig2017photo} and RCAN~\cite{zhang2018image}. However, their assumption of the fixed bicubic degradation leads to a severe performance drop to super-resolve the real-world LR images with complex and unknown degradations~\cite{liu2022blind,wei2021unsupervised}. To alleviate this problem, 
various blind image super-resolution (BISR) methods have been proposed to construct the real-world LR-HR data pairs~\cite{cai2019toward} via a simple combination of multiple degradations~\cite{bell2019blind,zhang2018learning},
alternative optimization~\cite{bell2019blind,gu2019blind,huang2020unfolding,luo2022learning,fu2022kxnet} and unsupervised contrastive learning~\cite{wang2021unsupervised,zhang2021blind}. For example, Cai \emph{et al.}~\cite{cai2019toward} constructed the RealSR dataset by collecting limited real-world LR-HR data pairs, which achieves limited performance improvement due to the incomplete definition of the image degradation space.  
Progressive estimation of blur kernel and restoration of SR images during the multi-stage restoration~\cite{gu2019blind,huang2020unfolding} significantly increase computation cost. 
%
Instead of a simple-combination degradation, several works consider more complex degradations by either random shuffle of degradation orders~\cite{zhang2021designing} or the second-order degradation~\cite{wang2021real,liang2022efficient} from the affection of camera blur, sensor noise, JPEG compression, and shapening artifacts.
Recently, with the development of Transformer architectures, Transfomer-based SR methods~\cite{chen2021pre,liang2021swinir,zamir2022restormer,zhang2022swinfir} have been proposed for image restoration by designing the effective Transformer structure. However, their performance improvement accompanied with the increasing of computation overhead and memory cost. Different from those, our method reduces the model redundancy by the proposed LOC block under the complex multi-level degradations.

\subsection{Efficient Single Image Super-Resolution}
Recently, various efficient SISR methods have been proposed to reduce the redundancy of SISR models, such as Neural Architecture Search (NAS)~\cite{chu2021fast,song2020efficient}, compact SR block design~\cite{ahn2018fast,song2021addersr,nie2021ghostsr}, quantization~\cite{li2020pams,hong2022cadyq} and dynamic patch-based inference~\cite{chen2022arm,hong2022cadyq,wang2022adaptive,kong2021classsr,zhong2022dynamic}, and knowledge distillation~\cite{gao2018image,lee2020learning,wang2021towards}. 
However, these methods may fail in the real-world image super-resolution, as they build on the assumption of bicubic degradation mismatch to the real-world complex degradation. 
Although Liang \emph{et al.}~\cite{liang2022efficient} adopted non-linear mixture of experts and proposed a degradation-adaptive framework to address the problems of complex degradation and efficiency, the mixture of multiple experts significantly increase the memory cost. Different from the multiple experts, we propose a novel dynamic channel splitting method to adaptively assign the channel rate to only one expert, which reduces the computation and parameter cost.



\subsection{Non-Local Regularization}

A natural image has a strong internal data repetition (\emph{e.g.}, image self-similarity), which can be regarded as the effective information for SISR~\cite{lotan2016needle,zontak2011internal,glasner2009super,shocher2018zero}. 
Non-local similarity is first proposed for image denoising~\cite{buades2005non,dabov2007image} by calculating the similarity between query patch and the remaining one around a given area. Recently, Non-local similarity has been merged into CNNs for image/video restoration. For example, CPNet \cite{li2021cross} extracts the similar patches for constructing cross-patch graph, which explicitly captures cross-patch long-range contextual dependency. 
%
Mei \emph{et al.} \cite{mei2020image} fuse in-scale non-local priors and cross-scale non-local priors, which come from the similarity between different patches at different scales for SISR. 
Non-local similarity can also be used for self-attention to generate the non-local features~\cite{davy2018non,liu2018non}, which significantly improves the performance in the low-level vision.
However, the existing non-local similarity approaches need to be computed at inference, which significantly increases the computation overhead and parameter cost for efficient inference. Different from those, the proposed non-local regularization constructs the rich knowledge from the relationship of the similar-pattern patches between LR and SR subspace, which is safely removed with free computation at inference. 
\section{Our Method}
\label{sec:Met}

\subsection{Notations and Preliminaries}

As shown in Fig.~\ref{fig_2}, our method consists three networks, including degradation prediction network $D$, channel splitting network $A$ and super-resolution network $f_{sr}$, with weights $\mathcal{W}_d$, $\mathcal{W}_a$ and $\mathcal{W}_{sr}$, respectively. Given an LR image $I_{lr}$, we first go through the network $D$ to generate the degradation vector $\hat{\mathbf{u}}\in\mathbf{R}^{d}$, which is further used to predict the channel splitting vector $\mathbf{a}\in\mathbf{R}^s$ by the network $A$. Finally, an SR image $I_{sr}$ is generated through the network $f_{sr}$ by taking $I_{lr}$ and $\mathbf{a}$ as the input. Mathematically, the above process can be formulated as:
\begin{equation}
I_{sr} = f_{sr}(I_{lr}, \mathbf{a}; \mathcal{W}_{sr}),
\end{equation}
where $\mathbf{a}=A(\hat{\mathbf{u}}; \mathcal{W}_a)$ and $\hat{\mathbf{u}}=D(I_{lr}; \mathcal{W}_d)$. To optimize the weights $\mathcal{W}_d$, $\mathcal{W}_a$ and $\mathcal{W}_{sr}$, we follow \cite{liang2022efficient} by the multi-level degradations to first construct the synthesized real-world pair between a given HR image $I_{hr}$ and a degradated LR image $I_{lr}$, as well as the degradation vector $\mathbf{u}$. 

\textbf{Construction of synthetic LR-HR pairs and degradation vector.} 
To simulate the real-world degradations, Recent works~\cite{wang2021real,liang2022efficient} proposed the second-order degradations to construct complex multi-level degradation pairs, where degradation operations involve blurring (both isotropic and anisotropic Gaussian blur), noise corruption (both additive Gaussian and Poisson noise), resizing (down-sampling and up-sampling with bilinear and bicubic operations), and JPEG compression. Inspired by these works, we also employ the multi-level degradation strategy to construct the training data, where a pair consists a given HR image $I_{hr}$, the corresponding multi-level degradated LR image $I_{lr}$ and the ground-truth degradation vector $\mathbf{u}$. In $\mathbf{u}$, we adopt a one-hot code to quantify the degradation operation type and record the degradation level normalized by its respective dynamic range. In this way, three-level degradation sub-spaces denoted by $S_i, (i=1,2,3)$ are constructed to generate the explicit degradation vector $\mathbf{u}$ with dimension of $d=33$, where $\mathbf{u}_i$ is a parameter in the corresponding degradation operation. 
Among them, $S_1$ and $S_2$ are generated with first-order degradation with small and large parameter ranges, respectively, and the second-order degradation $S_3$. We denote $S_0$ as the bicubic degradation. 
More details can be referred to \cite{liang2022efficient}. 

\textbf{Regression loss.} To learn the degradation prediction network $D$, we employ the regression loss via L1-norm between the estimated degradation vector $\hat{\mathbf{u}}$ and the ground-truth counterpart $\mathbf{u}$ as:
\vspace{-0.6em}
\begin{equation}
\label{reg}
\mathcal{L}_{reg} = \|\hat{\mathbf{u}}-\mathbf{u}\|_1.
\end{equation}
Subsequently, $\hat{\mathbf{u}}$ can be used as an input for channel splitting network $A$ to generate the channel splitting vector $\mathbf{a}$. In the following parts of Sec.~\ref{sec:Met}, we will introduce the learning of weights $\mathcal{W}_a$ and $\mathcal{W}_{sr}$ for efficient content-dependent SR. 

\begin{figure}[t]
  \centering
   \includegraphics[width=1.0\linewidth]{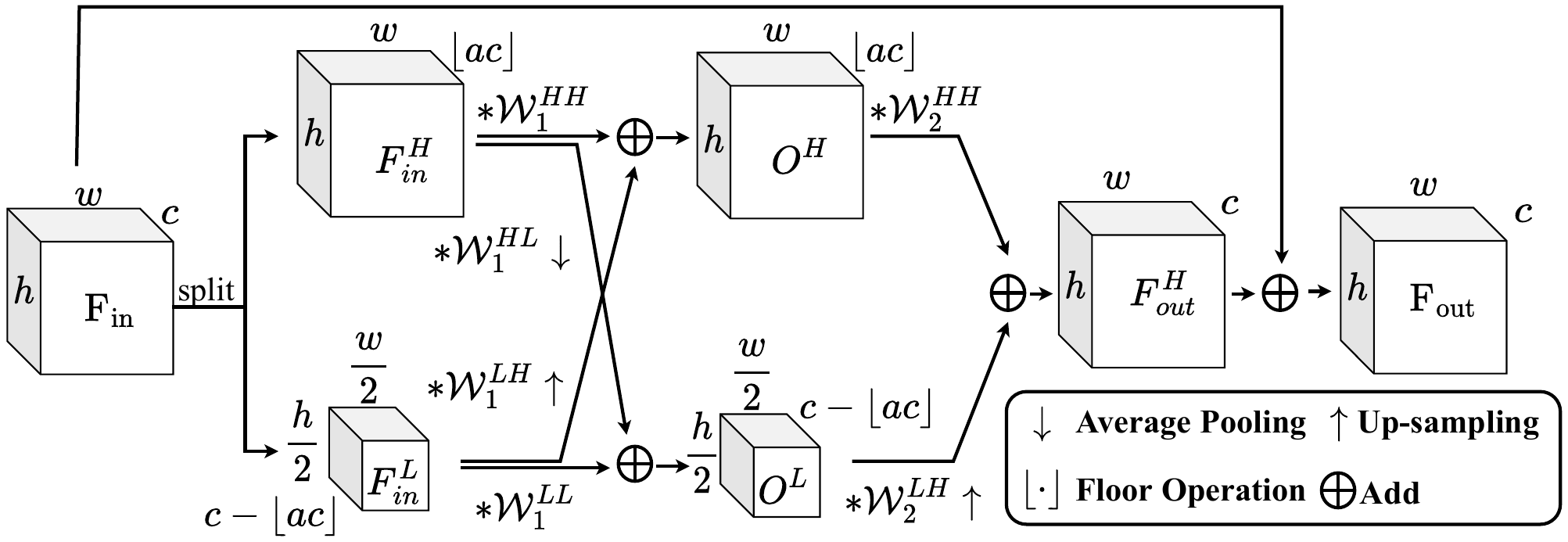}
   \caption{An illustration of LOC block.}
   \label{fig:Loc}
\end{figure}

\subsection{Learnable Octave Convolution Block}

As discussed in Sec.~\ref{sec:intro}, few SR methods consider to effectively handle both efficiency and complex degradations. To alleviate the problem, we leverage the multi-level degradation strategy into more compact SR block to improve efficient real-world SR performance. Consider 
most SR backbone networks~\cite{zhang2018image,ledig2017photo} consists of three parts: head, trunk and reconstruction, we can formulate the SR output as:
\begin{equation}
\small
I_{sr} = f_{sr-r}\circ f_{sr-t}^{N}\circ\cdots\circ f_{sr-t}^{1}\circ f_{sr-h}(I_{lr}),
\end{equation}
where $f_{sr-h}$ and $f_{sr-r}$ are the transformation of head and reconstruction parts with parameters $\mathcal{W}_{sr}^{h}$ and $\mathcal{W}_{sr}^{r}$, respectively. $f_{sr-t}^{i}, i=1,\cdots N$ is the $i$-th block with parameter $\mathcal{W}_{sr-t}^{i}$ in the trunk part, which is the key part by stacking multiple residual blocks~\cite{ledig2017photo} or their variants~\cite{zhang2018image,dai2019second} to extract rich feature information for reconstruction but accompanied with significant computation overhead and memory cost. OctConv~\cite{chen2019drop} assumes that the input feature maps can be factorized into high- and low-frequency features stored into two groups, and reduce the spatial resolution of the low-frequency group for convolutional computation to remove the redundancy. However, how to decide the channel ratio of these two-frequency features for better channel splitting is unexploited, and build the relationship between the channel ratio and image content to process images with various kinds of degradations. To this end, we propose a learnable octave convolution block as our main component in the SR network.
%



In our efficient SR network, we keep the head and reconstruction part unchanged to off-the-shelf SR models~\cite{ledig2017photo,zhang2018image,wang2018esrgan}, due to their small computation overhead and useful information. We only replace the residual block with the proposed LOC block in the trunk part. Inspired by \cite{chen2019drop}, we leverage each channel splitting factor $a_i$ into the $i$-th LOC block and merge two frequency outputs to form the entire feature maps at the end of the $i$-th LOC block.

For better description, we take one LOC block with two convolutions for discussion and remove the superscription block index. We first decompose the input feature maps $F_{in}$ into high-frequency feature $F_{in}^{H}\in\mathbf{R}^{\lfloor a\cdot c \rfloor \times h \times w}$ and $F_{in}^{L}\in\mathbf{R}^{ (c-\lfloor a\cdot c \rfloor) \times \frac{h}{2} \times \frac{w}{2}}$ along the channel dimension according to the predicted channel splitting channel $a$, where $\lfloor\cdot\rfloor$ is the floor operation, $F_{in}^{L}$ is the bilinear downsampling from $F_{in}$, and 
$a\in[0,1]$ denotes the channel ratio allocated to the high-frequency part. $h,w$ and $c$ represent the height, width and channel number, respectively. 
The high-frequency feature $F_{in}^{H}$ generates edge and texture information, and the low-frequency one $F_{in}^{L}$ captures the information that the gray value vary slower in the spatial dimensions. 
As shown in Fig.~\ref{fig:Loc}, the first convolution kernel $\mathcal{W}_{sr-t}^{(1)}\in\mathbf{R}^{c\times c\times k\times k}$ in the block $f_{sr-t}$ is split into $\mathcal{W}_1^{HH}\in\mathbf{R}^{\lfloor a\cdot c \rfloor\times \lfloor a\cdot c \rfloor\times k\times k}, \mathcal{W}_1^{LL}\in\mathbf{R}^{(c-\lfloor a\cdot c \rfloor)\times (c-\lfloor a \cdot c \rfloor)\times k\times k}, \mathcal{W}_1^{HL}\in\mathbf{R}^{ (c-\lfloor a\cdot c \rfloor) \times \lfloor a\cdot c \rfloor\times k\times k}$, and $\mathcal{W}_1^{LH}\in\mathbf{R}^{\lfloor a\cdot c \rfloor \times (c-\lfloor a\cdot c \rfloor) \times k\times k}$ to compute intra-frequency and inter-frequency communication. Thus, the high-frequency output feature $O^{H}$ and the low-frequency output one $O^{L}$ can be computed as:
\begin{equation}
\small
\label{eq_o}
\begin{split}
&O^{H} = ReLU(F_{in}^{H}*\mathcal{W}_1^{HH} + \text{upsample}(F_{in}^{L}*\mathcal{W}_1^{LH}, 2)), \\
&O^{L} = ReLU(F_{in}^{L}*\mathcal{W}_1^{LL} + \text{pool}(F_{in}^{H},2)*\mathcal{W}_1^{HL}),
\end{split}
\end{equation}
where $*$, upsample and pool denote the operations of convolution, up-sampling and average pooling, respectively. $ReLU(\cdot)$ is the non-linear activation. In the second convolution, $\mathcal{W}_2^{HH}\in\mathbf{R}^{ c \times \lfloor a\cdot c \rfloor \times k\times k}$ and $\mathcal{W}_2^{LH}\in\mathbf{R}^{c \times (c-\lfloor a\cdot c \rfloor) \times k\times k}$ are employed to merge the outputs of high- and low-frequency features and generate the block output $F_{out}$, which can be formulated as:
\begin{equation}
\footnotesize
\label{eq_f}
\begin{split}
&F_{out}^{H} = ReLU\bigg(\big(O^{H}*\mathcal{W}_2^{HH} + \text{upsample}(O^{L}*\mathcal{W}_2^{LH}, 2)\big)\bigg), \\
&F_{out} = F_{in} + F_{out}^{H}, 
\end{split}
\end{equation}
%
where $[\cdot]$ is the concatenation operation on the channel dimension.  Instead of the traditional residual block $f_{sr-t}$, we employ the learnable LOC block to obtain the SR image $I_{sr}$, and use the pixel-wise reconstruction loss to learn their parameters as:
\vspace{-0.2em}
\begin{equation}
\label{pix}
\mathcal{L}_{pix} = \|I_{sr} - I_{hr}\|_1,
\end{equation}
where $\|\cdot\|_1$ is the L1-norm and $I_{hr}$ is the GT HR image.

The channel splitting scale $a\in\mathbf{a}$ plays an important role of controlling the efficiency of LOC block in Eqs.~\ref{eq_o} and ~\ref{eq_f}, which will be degraded into the traditional residual block if setting $\alpha$ to 1. Moreover, consider that different image inputs may learn different high- and low-frequency features at the different depths, it is not reasonable to set $\mathbf{a}$ to be fixed. Thus, 
we propose a learnable $\mathbf{a}\in\mathbf{R}^s$ ($s$ is the block number) to dynamically adapt to different blocks, which relies on the content of LR input $I_{in}$. In particular, we directly calculate $\mathbf{a}$ conditioned on the predicted degradation vector $\hat{\mathbf{u}}$ via a tiny network $A$ (\emph{i.e.}, $\mathbf{a}=A(\hat{\mathbf{u}}; \mathcal{W}_a)$). To reduce the model redundancy, the sparsity regularizer for $\mathbf{a}$ is introduced to constrain the usage of high-frequency information, which is formulated as:
\vspace{-0.2em}
\begin{equation}
\label{loss_a}
\mathcal{L}_{a} = \|\mathbf{a}\|_1.
\end{equation}



\begin{figure}[t]
  \centering
   \includegraphics[width=0.8\linewidth]{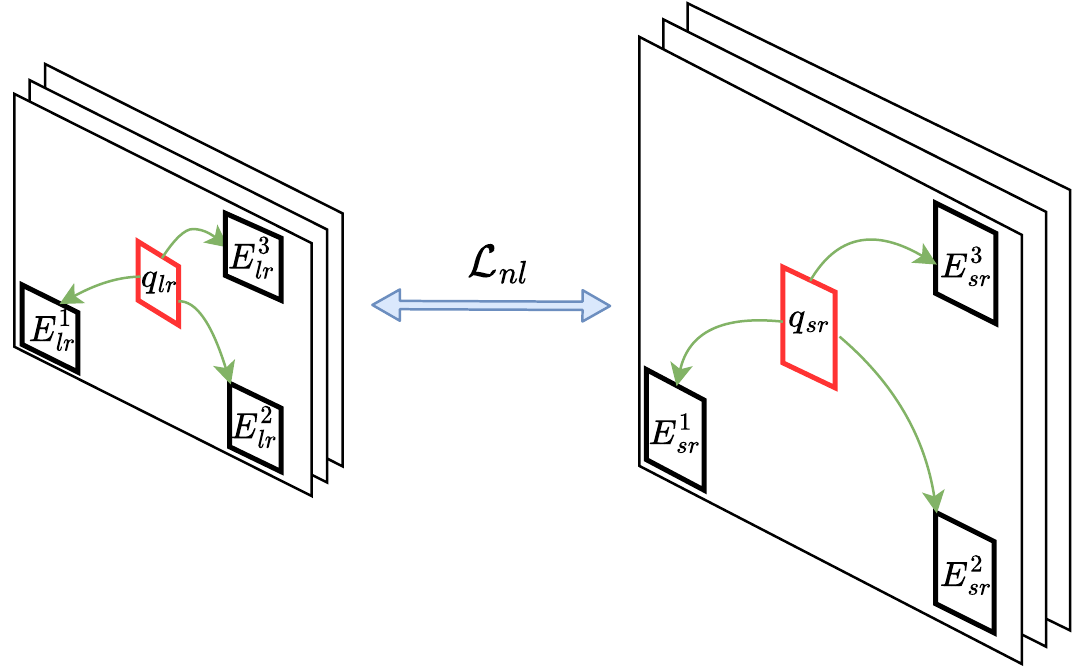}
   \caption{An illustration of non-local regularization with $K=3$.}
   \label{fig:onecol}
\end{figure}

\subsection{Non-local Regularization}

Pixel-wise reconstruction loss Eq.~\ref{pix} only considers the information of independent patch to align the LR image and its HR counterpart from the inter-patch. We assume the rich knowledge from intra-patch is also important for SR, which can be used to further improve the SR performance. The patch relationship can be well built by the non-local similarity~\cite{buades2005non,dabov2007image,lotan2016needle,li2021cross,liu2018non,davy2018non}, which has wide applications on the image or video restoration task. However, the non-local information is often calculated for inference, which significantly increases the computation overhead. To this end, we propose a non-local regularization to build the relationship of the similar-pattern on the intra-patch between LR and SR subspace, which improves model performance while safely being removed for inference.


As shown in Fig.~\ref{fig:onecol}, given a small query patch $q_{lr}$ from the LR patch, we first search the $K$-similar patches to the query $q_{lr}$ based on non-local similarity computation (\emph{e.g.}, Euclidean distance) in the LR space, except itself. The most $K$-similar patch set is constructed as $E_{lr}=\{E_{lr}^1, E_{lr}^2, \cdots, E_{lr}^K\}$ in a descending order. Correspondingly, we can construct the query patch $q_{sr}$ and the patch set $E_{sr}=\{E_{sr}^1, E_{sr}^2, \cdots, E_{sr}^K\}$ in the SR space via our SR model $f_{sr}$ according to the corresponding positions of $q_{lr}$ and $E_{lr}^i, i=1,\cdots K$, respectively. Finally, we employ the similarity between query and similar patch set to construct the non-local regularization loss, which is formulated as:
\begin{equation}
\label{nl}
\mathcal{L}_{nl} = \frac{1}{K}\|\sum_{i=0}^{K}\Delta(q_{lr}, E_{lr}^i) - \sum_{i=0}^{K}\Delta(q_{sr}, E_{sr}^i)\|_2,
\end{equation}
where $\Delta(\cdot,\cdot)$ is the absolute value of Euclidean distance. $E_{lr}^0$ and $E_{sr}^0$ are both set to be $\mathbf{0}$, which means that $\mathcal{L}_{nl}$ is only computed on the query patch when setting $K$ to 0. 
The larger $K$ will increase the training computation overhead, but not for inference. This is due to the safely removal of Eq.~\ref{nl} at inference. We will discuss the setting of the hyper-parameter $K$ in experiments. 
%



\begin{figure*}[t]
\centering
\begin{minipage}{0.18\linewidth}
\centerline{\includegraphics[width=\textwidth]{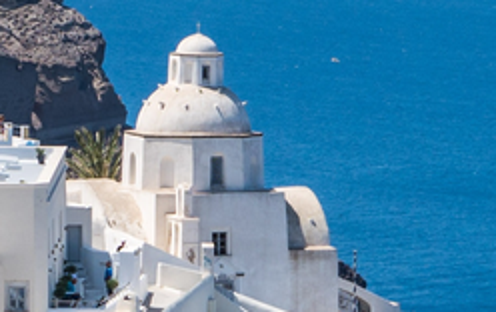}}
\vspace{3pt}
\centerline{\small (a) HR}
\end{minipage}
\begin{minipage}{0.18\linewidth}
\centerline{\includegraphics[width=\textwidth]{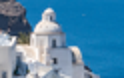}}
\vspace{3pt}
\centerline{\small (b) Bicubic}
\end{minipage}
\begin{minipage}{0.18\linewidth}
\centerline{\includegraphics[width=\linewidth]{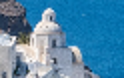}}
\vspace{3pt}
\centerline{\small (c) Level 1}
\end{minipage}
\begin{minipage}{0.18\linewidth}
\centerline{\includegraphics[width=\textwidth]{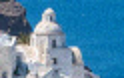}}
\vspace{3pt}
\centerline{\small (d) Level 2}
\end{minipage}
\begin{minipage}{0.18\linewidth}
\centerline{\includegraphics[width=\textwidth]{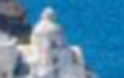}}
\vspace{3pt}
\centerline{\small (e) Level 3}
\end{minipage}
\vspace{-.5em}
\caption{Sample images from DIV2K validation set with different degradation levels.}
\label{fig_sample}
\end{figure*}

\begin{table*}[t]
\centering
\caption{Quantitative results of $\times$4 on DIV2K validation set with different degradation levels. \textbf{Bold} and \underline{underline} number are the best and ranking-second best performance in all tables. GFLOPs and parameters are evaluated on an average image with $256\times 256$ pixels.}
\vspace{-0.5em}
\resizebox{.9\linewidth}{!}{
\small
\begin{tabular}{ccccccccccc}
\toprule
\multirow{2}*{Method} & \multicolumn{2}{c}{Level-0 (bicubic)} & \multicolumn{2}{c}{Level-1} & \multicolumn{2}{c}{Level-2} & \multicolumn{2}{c}{Level-3} & \multirow{2}*{GFLOPs} & \multirow{2}*{Param.(M)}\\
&PSNR&SSIM&PSNR&SSIM&PSNR&SSIM&PSNR&SSIM& & \\
\midrule
SRResNet~\cite{ledig2017photo} & 28.05 &0.7702 & 27.60 &0.7526 & \underline{27.34} &0.7211 & 23.65 &0.6012 & 166 & \textbf{1.52} \\
EDSR~\cite{lim2017enhanced} &28.25&0.7936&27.47&0.7712&27.28&\underline{0.7580}&23.47&0.6077 & \underline{130} & \textbf{1.52} \\
RRDB~\cite{wang2018esrgan} & \textbf{30.92} & \textbf{0.8486} & 26.27 & 0.7578 & 26.46 & 0.7060 & 23.79 & 0.5975 & 1,176 & 16.70 \\
ESRGAN~\cite{wang2018esrgan} & 28.17 & 0.7759 & 21.16 & 0.4515 & 22.77 & 0.4746 & 23.49 & 0.5706 & 1,176 & 16.70 \\
DAN~\cite{huang2020unfolding} & \underline{30.51} & \underline{0.8411} & 27.34 & \underline{0.7803} & 27.15 & 0.7298 & 23.74 & 0.5928 & 1,177 &4.04\\
BSRGAN~\cite{zhang2021designing} & 27.32 & 0.7577 & 26.78 & 0.7453 & 26.75 & 0.7370 & 23.84 & 0.6235 & 1,176 & 16.70 \\
Real-ESRGAN~\cite{wang2021real} & 26.64 & 0.7581 & 26.16 & 0.7470 & 26.16 & 0.7413 & 23.59 & \textbf{0.6349} & 1,176 & 16.70 \\
Real-SwinIR-M~\cite{liang2021swinir} & 26.83 & 0.7649 & 26.21 & 0.7481 & 26.11 & 0.7389 & 23.19 & 0.6216 & 841 & 11.64 \\
Real-SwinIR-L~\cite{liang2021swinir} & 27.20 & 0.7758 & 26.44 & 0.7561 & 26.38 & 0.7469 & 23.31 & 0.6241 & 1,901 & 27.86 \\
DASR~\cite{wang2021unsupervised}&30.26 &0.8362 &26.75 &0.7621 &26.75 &0.7131 &23.48 &0.5765 &209 &5.8\\
CARN~\cite{ahn2018fast} &30.43 &0.8374 &26.62 &0.7622 &26.59 &0.7081 &23.81 &0.6002 &103&1.59 \\
KXNet~\cite{fu2022kxnet}& 24.54 & 0.7127 & 24.84 & 0.707 & 25.26 & 0.7028 & 23.34 & 0.5877 &408 & 6.5\\
Liang \emph{et al.}~\cite{liang2022efficient} & 28.55 & 0.7983 & \underline{27.84} & 0.7772 & \textbf{27.58} & \textbf{0.7652} & \underline{23.93} & \underline{0.6250} & 184 & 8.07 \\ \hline
DCS-RISR & 29.01 & 0.7970 & \textbf{27.89} & \textbf{0.7828} & 27.29 & 0.7404 & \textbf{23.95} & 0.6156 & \textbf{128} & \underline{1.55} \\
\bottomrule
\end{tabular}}
\label{tab:EDSR}
\end{table*}

\subsection{The Overall Loss and Its Solver}

Our DCS-RISR contains three parameter weights, $\mathcal{W}_d$, $\mathcal{W}_a$ and $\mathcal{W}_{sr}$. To optimize these weights and implement efficient inference, we follow the previous works \cite{wang2018esrgan,zhang2021designing,wang2021real,liang2022efficient} by leveraging the perceptual loss $\mathcal{L}_{per}$, the adversarial loss $\mathcal{L}_{adv}$, the regression loss $\mathcal{L}_{reg}$ (Eq~\ref{reg}), two regularizer losses $\mathcal{L}_{a}$ and $\mathcal{L}_{nl}$ (Eq.~\ref{loss_a} to~\ref{nl}) into the reconstruction loss $\mathcal{L}_{pix}$ (Eq.~\ref{pix}) to construct the overall loss of our DCS-RISR scheme, which can be formulated as:
\begin{equation}
\small
\label{overall}
\mathcal{L}_{o} = \mathcal{L}_{pix} + \lambda_1\mathcal{L}_{reg} + \lambda_2\mathcal{L}_{per} + \lambda_3\mathcal{L}_{adv} + \lambda_4\mathcal{L}_{nl} + \lambda_5\mathcal{L}_{a},
\end{equation}
where $\lambda_i,i=1,\cdots,5$ are the balancing parameters, especially $\lambda_5$ controls the efficiency of our SR model. 

\textbf{Solver.} Since three networks are used for training to dynamically learn $\mathbf{a}$ and our efficient SR model parameter $\mathcal{W}_{sr}$, direct SGD methods (\emph{e.g.}, Adam optimizer~\cite{kingma2014adam}) are used to minimize Eq.~\ref{overall}, which may lead to the unstable training. As a solution, we employ two-stage training by first pre-training our efficient SR backbone with the fixed $\mathbf{a}=\textbf{0.5}$ by only minimizing the reconstruction loss Eq.~\ref{pix}, and then re-training the networks $D$, $A$ and $f_{sr}$ simultaneously to minimize Eq.~\ref{overall}.

\begin{table*}[t]
\centering
\footnotesize
\caption{Quantitative results on Set5, Set14, B100 and Urban100 from the degradation level $S_1$-$S_3$. We report the average PSNR metric. L1, L2 and L3 mean Level-1, Level-2 and Level-3, respectively.}
\vspace{-0.5em}
\begin{tabular}{|cl|c|lll|lll|lll|lll|}
\hline
\multicolumn{2}{|c|}{\multirow{2}{*}{Method}} & \multirow{2}{*}{scale} & \multicolumn{3}{c|}{Set5}                                                                  & \multicolumn{3}{c|}{Set14}                                                                 & \multicolumn{3}{c|}{B100}                                                                  & \multicolumn{3}{c|}{Urban100}                                                              \\ \cline{4-15} 
\multicolumn{2}{|c|}{}                        &                        & \multicolumn{1}{c|}{L1} & \multicolumn{1}{c|}{L2} & \multicolumn{1}{c|}{L3} & \multicolumn{1}{c|}{L1} & \multicolumn{1}{c|}{L2} & \multicolumn{1}{c|}{L3} & \multicolumn{1}{c|}{L1} & \multicolumn{1}{c|}{L2} & \multicolumn{1}{c|}{L3} & \multicolumn{1}{c|}{L1} & \multicolumn{1}{c|}{L2} & \multicolumn{1}{c|}{L3} \\ \hline
\multicolumn{2}{|c|}{SRResNet~\cite{ledig2017photo}}  & \multirow{7}{*}{×2}        & 33.77               & 29.90                  & 22.89                  & 30.60                  & 27.17                  & 22.41                  & 29.17                  & 26.32                  & 23.30                  & 27.43                  & 24.99                  & 20.95                  \\ \cline{1-2}
\multicolumn{2}{|c|}{RCAN~\cite{zhang2018learning}}  &    &\textbf{33.93} &	29.96 	&22.89 &	30.63 &	27.19 &	22.40 	&29.20 &	26.32 	&\underline{23.31} 	&27.34 &	25.04& 	20.96 
\\ \cline{1-2}
\multicolumn{2}{|c|}{IKC~\cite{gu2019blind}}      &                            & \underline{33.80}         & 30.34                  & 22.82                  & \textbf{31.01}                  & \underline{27.68}                  & 22.22                  & \textbf{29.56}         & \underline{26.80}                  & 23.24                  & \textbf{27.94}         & \underline{25.50}                  & 20.95                  \\ \cline{1-2}
\multicolumn{2}{|c|}{DAN~\cite{huang2020unfolding}}      &                            & 33.76                  & 30.07                  & 22.79                  & \underline{30.89}         & 27.33                  & 22.35                  & \underline{29.42}            & 26.44                  & 23.16                  & 27.75                  & 25.15                  & 20.94                  \\ \cline{1-2}
\multicolumn{2}{|c|}{DASR~\cite{wang2021unsupervised}}     &                            & 31.49                  &\underline{30.68}            & 23.53                  & 30.84            & 27.30                  & 22.26                  & 29.36                  & 26.43                  & 23.17                  & 27.68                  & 25.06                  & 20.91                  \\ \cline{1-2}
\multicolumn{2}{|c|}{CARN~\cite{ahn2018fast}}  &    &33.71 &	29.89 &	22.88 &	30.62& 	27.18& 	22.42& 	29.17 &	26.32 &	\underline{23.30} &	27.44 &	24.98 &	20.96 
 \\ \cline{1-2}
\multicolumn{2}{|c|}{KXNet~\cite{fu2022kxnet}}  &                            & 30.21                  & 29.61                  & \textbf{23.80}                  & 27.95                  & 27.11                  & \underline{22.51}                  & 27.86                  & 26.66                  & 23.28                  & 25.43                  & 24.64                  & \underline{21.01}                  \\ \cline{1-2}
\multicolumn{2}{|c|}{DCS-RISR} &                            & 33.13            & \textbf{30.83}         & \underline{23.57}         & 30.63                  & \textbf{28.25}         & \textbf{23.20}         & 29.33                  & \textbf{27.36}         & \textbf{23.77}         & \underline{27.91}            & \textbf{25.90}         & \textbf{21.35}         \\ \hline
\multicolumn{2}{|c|}{SRResNet~\cite{ledig2017photo}}  & \multirow{8}{*}{×4}        & 28.71                  & 25.92                  & 22.51                  & 24.39                  & 24.31                  & 22.17                  & 24.31                & 24.62                  & 22.37                  & 21.93                  & 22.03                  & 19.98                  \\ \cline{1-2}
\multicolumn{2}{|c|}{RCAN~\cite{zhang2018learning}}  &    &28.17 &	25.57 &	22.47 &	23.76 &	24.04 &	22.14 &	24.06 &	24.53 &	22.30 &	21.45 &	21.81 &	19.93 
\\ \cline{1-2}
\multicolumn{2}{|c|}{IKC~\cite{gu2019blind}}      &                            &\underline{29.33}            & 26.56                  & 22.13                  & 24.81                  & 24.85                  & 22.01                  & 24.49                  & 24.81                  & 22.19                  & 22.09                  & 22.21                  & 19.93                  \\ \cline{1-2}
\multicolumn{2}{|c|}{DAN~\cite{huang2020unfolding}}      &                            & \textbf{29.54}         & \underline{27.07}            & \textbf{22.78}         & \underline{25.45}                  & \underline{25.14}            & \underline{22.21}            &\underline{25.06}            &\underline{25.17}            & 22.24            & \underline{22.81}                  & \underline{22.75}            & \underline{20.01}                  \\ \cline{1-2}
\multicolumn{2}{|c|}{DASR~\cite{wang2021unsupervised}}     &                            & 29.01                  & 26.35                  & \underline{22.75}            & 24.86            & 24.66                  & 22.09                  & 24.63                  & 24.84                  & 22.04                  & 22.46                  & 22.42                  & 19.91                  \\ \cline{1-2}
\multicolumn{2}{|c|}{CARN~\cite{ahn2018fast}}  &    &28.75 &	25.93 &	22.57 &	24.52 &	24.42 &	22.18 &	24.40 &	24.67 &	\underline{22.38} &	22.06 &	22.14 &	19.99 
\\ \cline{1-2}
\multicolumn{2}{|c|}{ARM~\cite{chen2022arm}}      &                            & 26.76                  & 24.11                  & 20.27                  & 23.18                  & 22.86                  & 20.06                  & 23.38                  & 23.41                  & 20.75                  & 20.75                  & 20.61                  & 18.21                  \\ \cline{1-2}
\multicolumn{2}{|c|}{KXNet~\cite{fu2022kxnet}}  &                            & 24.19                  & 24.28                  & 22.33                  & 22.78                  & 23.47                  & 21.76                  & 23.43                  & 23.69                  & 22.13                  & 20.32                  & 20.89                  & 19.77                  \\ \cline{1-2}
\multicolumn{2}{|c|}{DCS-RISR} &                            & 29.26                  & \textbf{28.03}         & 22.49                  & \textbf{26.17}         & \textbf{25.73}         & \textbf{22.22}         & \textbf{25.59}         & \textbf{25.52}         & \textbf{22.52}         & \textbf{23.50}         & \textbf{23.30}         & \textbf{20.04}            \\ \hline
\end{tabular}
\label{tab_set5}
\end{table*}

\begin{figure*}[t]
\centering
\begin{minipage}{0.3\linewidth}
\centerline{\includegraphics[width=\textwidth]{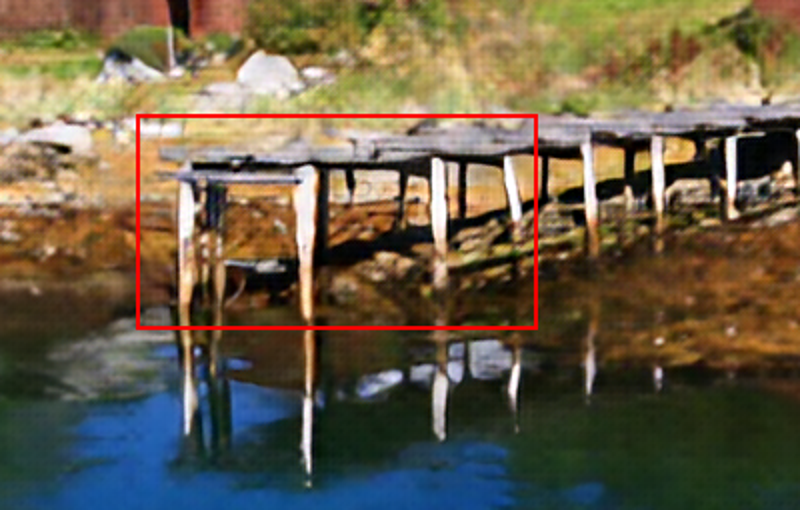}}
\centerline{\small (1) Level 1 0808.png}
\vspace{2pt}
\centerline{\includegraphics[width=\textwidth]{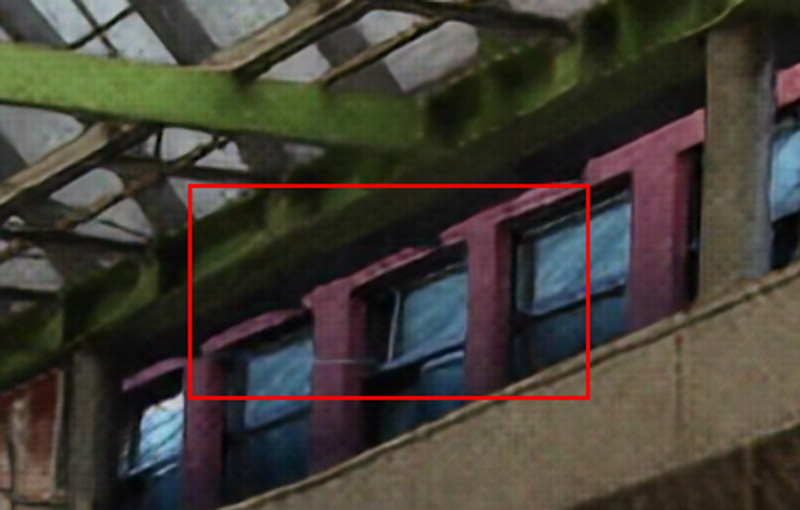}}
\centerline{\small (2) Level 2 0831.png}
\vspace{2pt}
\end{minipage}
\begin{minipage}{0.149\linewidth}
\centerline{\includegraphics[width=\textwidth]{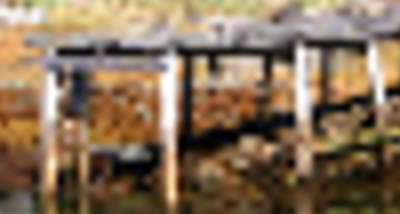}}
\centerline{\small (a) Bicubic}
\vspace{2pt}
\centerline{\includegraphics[width=\textwidth]{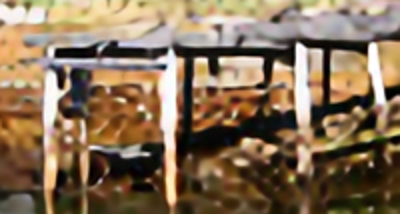}}
\centerline{\small (e) DAN}
\vspace{2pt}
\centerline{\includegraphics[width=\textwidth]{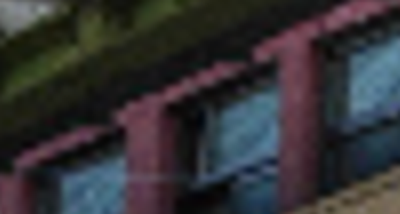}}
\centerline{\small (a) Bicubic}
\vspace{2pt}
\centerline{\includegraphics[width=\textwidth]{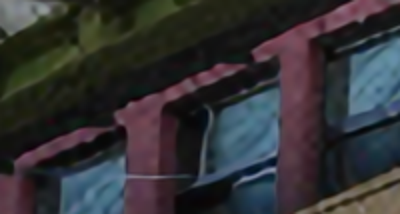}}
\centerline{\small (e) DAN}
\vspace{2pt}
\end{minipage}
\begin{minipage}{0.149\linewidth}
\centerline{\includegraphics[width=\textwidth]{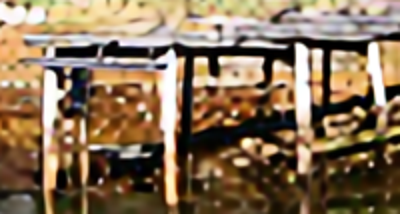}}
\centerline{\small (b) RRDB}
\vspace{2pt}
\centerline{\includegraphics[width=\textwidth]{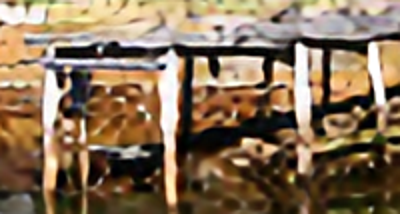}}
\centerline{\small (f) DASR}
\vspace{2pt}
\centerline{\includegraphics[width=\textwidth]{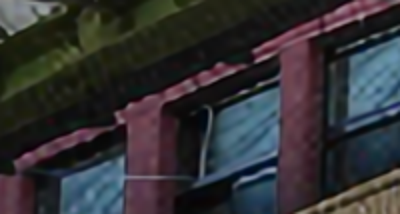}}
\centerline{\small (b) RRDB}
\vspace{2pt}
\centerline{\includegraphics[width=\textwidth]{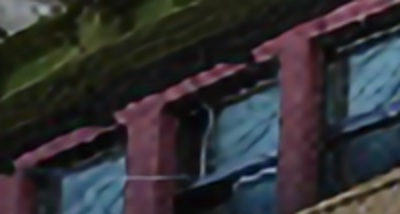}}
\centerline{\small (f) DASR}
\vspace{2pt}
\end{minipage}
\begin{minipage}{0.149\linewidth}
\centerline{\includegraphics[width=\textwidth]{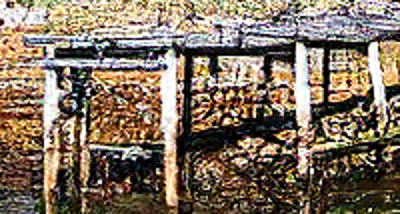}}
\centerline{\small (c) ESRGAN}
\vspace{2pt}
\centerline{\includegraphics[width=\textwidth]{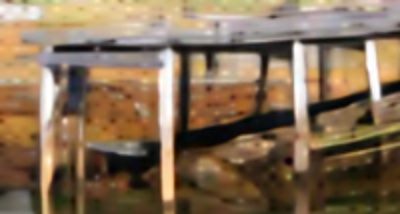}}
\centerline{\small (g) KXNet}
\vspace{2pt}
\centerline{\includegraphics[width=\textwidth]{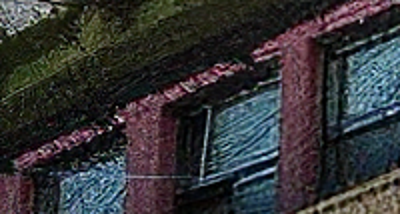}}
\centerline{\small (c) ESRGAN}
\vspace{2pt}
\centerline{\includegraphics[width=\textwidth]{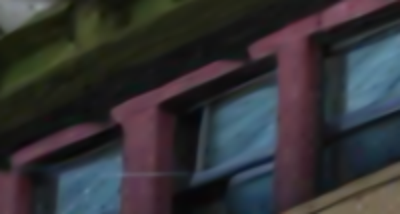}}
\centerline{\small (g) KXNet}
\vspace{2pt}
\end{minipage}
\begin{minipage}{0.149\linewidth}
\centerline{\includegraphics[width=\textwidth]{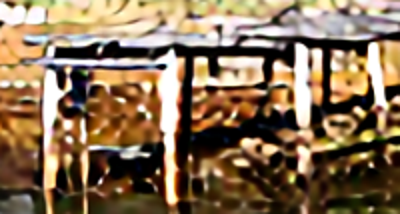}}
\centerline{\small (d) IKC}
\vspace{2pt}
\centerline{\includegraphics[width=\textwidth]{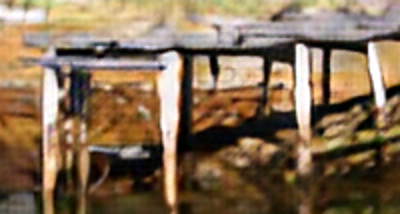}}
\centerline{\small (h) DCS-RISR}
\vspace{2pt}
\centerline{\includegraphics[width=\textwidth]{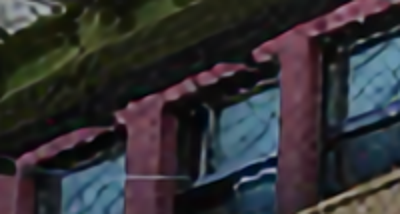}}
\centerline{\small (d) IKC}
\vspace{2pt}
\centerline{\includegraphics[width=\textwidth]{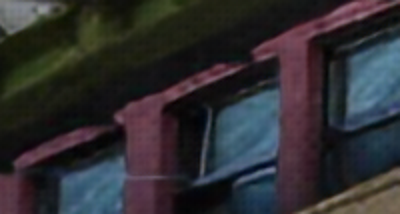}}
\centerline{\small (h) DCS-RISR}
\vspace{2pt}
\end{minipage}
\vspace{-0.5em}
\caption{Qualitative Comparison at different degradation levels of $\times$4 on DIV2K validation set. Zoom in for best views.}
\label{fig_6}
\end{figure*}

\section{Experiments}
\subsection{Experimental Settings}
\textbf{Datasets.} 
We follow the previous works~\cite{ledig2017photo,wang2021real,wang2018esrgan,liang2022efficient}, DIV2K~\cite{agustsson2017ntire}, Flickr2K~\cite{timofte2017ntire} and OST~ \cite{wang2018recovering} datasets are used to train our model. To evaluate the effectiveness of the proposed DCS-RISR, we synthesize 300 LR-HR pairs by applying the three degradation levels from $S_1$ to $S_3$ into the 100 validation images in the DIV2K dataset. Each level has 100 LR-HR pairs. The original 100 images degradated by bicubic downsampling (\emph{i.e.}, $S_0$) on DIV2K are also used for evaluation. By this way, we additionally test on four SR benchmarks from the degradation of $S_1$ to $S_3$: Set5~\cite{bevilacqua2012low}, Set14~\cite{zeyde2010single}, BSD100~\cite{martin2001database}, and Urban100~\cite{huang2015single}. Fig.~\ref{fig_sample} shows an image with different degradation levels, and Supplimentaries show more examples. 

\textbf{Implementation details.}
Our models are implemented by PyTorch 1.8 with total 1300K iterations (1000K for pre-training on DIV2K) on one NVIDIA 3090 GPU. 
The models are optimized by Adam optimizer \cite{kingma2014adam} with $\beta_1=0.9, \beta_2=0.999$, and $\epsilon=10^{-8}$.
The batch size is set to 24. For pre-training, we set the initial learning to $10^{-4}$ with the decay rate of 10 at every 250K iteration. For re-training, the initial learning is fixed by $10^{-6}$.
we set $\lambda_1:\lambda_2:\lambda_3:\lambda_4:\lambda_5$ to 1:1:0.1:1:0.25 for balancing the training losses and $K$ to 3, unless otherwise specified. More detailed settings for these balancing parameters are discussed in supplementaries.
The HR patch size and query patch size are set to 256$\times$256 and 16$\times$16, respectively. 
For the DCS-RISR backbone architecture, we choose the SRResNet with LOC blocks. 
We calculate PSNR and SSIM on the Y channel, GFLOPs and parameters.

\textbf{Baselines and SOTA methods.} we set the SRResNet \cite{ledig2017photo} as our baseline, and make comparisons with the previous SOTA methods, such as EDSR~\cite{lim2017enhanced}, RCAN \cite{zhang2018image}, CARN\cite{ahn2018fast},  RRDB \cite{wang2018esrgan}, ESRGAN \cite{wang2018esrgan}, IKC\cite{gu2019blind}, DAN\cite{huang2020unfolding}, BSRGAN \cite{zhang2021designing}, Real-ESRGAN \cite{wang2021real}, Real-SwinIR \cite{liang2021swinir}, Liang \emph{et al.} \cite{liang2022efficient} and KXNet\cite{fu2022kxnet}. 

\textbf{Network architectures of $D$ and $A$.}
We adopt a light-weight network $D$, which is composed of 3 convolution layers (channels: 64-33-33) with leaky ReLU activation and adds a global average pooling layer and one fully-connected (FC) layer to generate the 33-dimension degradation vector. The network $A$ consist of two FC layers and Sigmoid activation with the neurons of 25-16. Note that the parameters of $A$ is negligible, and network $D$ only has 0.14M parameters, accounting for $\sim$10\% of SRResNet (1.52M).

\begin{figure}[t]
\begin{minipage}{0.48\linewidth}
\vspace{3pt}
\centerline{\includegraphics[width=\textwidth]{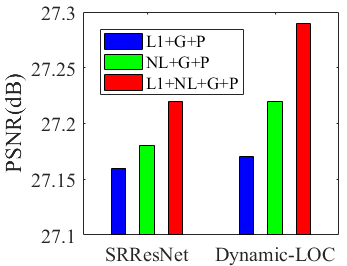}}
\centerline{\small (a) PSNR}
\end{minipage}
\begin{minipage}{0.48\linewidth}
\centerline{\includegraphics[width=\textwidth]{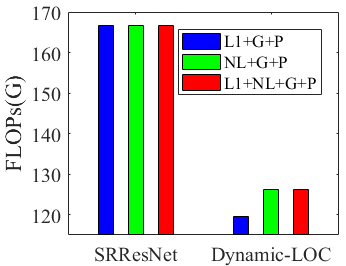}}
\centerline{\small (b) FLOPs}
\end{minipage}
\vspace{-0.5em}
\caption{Comparison with/without non-local regularization on SRResNet and its dynamic structure. L1, NL, G and P mean L1 reconstruction, non-local, GAN and perceptual losses, respectively.}
\label{fig_nl}
\end{figure}


\subsection{Comparisons with Prior SOTA methods}
%
\textbf{Quantitative evaluation.} Tab.~\ref{tab:EDSR} summarizes the quantitative results on DIV2K validation set at the SR scale of $\times$4. 
We observe that RRDB~\cite{wang2018esrgan} achieves the highest PSNR and SSIM of 30.92db and 0.8486 at the bicubic degradation, but increasing the number of degradation levels leads to the significant performance drop. Moreover, the computation and memory cost in RRDB are heavy that attaining to 1,176 GFLOPs and 16.7M parameters.
Compared to RRDB, DAN~\cite{huang2020unfolding} achieves the higher PSNR/SSIM under the more complex degradation, which benefits from progressive updating between the blur kernel and intermediate SR images. Although parameters can be reduced by parameter sharing, the multi-stage inference increases the computation overhead. 
%
Compared to SRResNet\cite{ledig2017photo}, RRDB and ESRGAN\cite{wang2018esrgan}, BSRGAN~\cite{zhang2021designing} and Real-ESRGAN~\cite{wang2021real} for RISR perform better when dealing with severely image degradation, while the performance in the small level degradations (\emph{e.g.}, bicubic and level-1) decreases significantly.
Real-SwinIR-L~\cite{liang2021swinir} employs the Swin-transformer architecture for super-resolution, which requires large number of parameters (\emph{i.e.}, 27.86M) and heavy computation (\emph{i.e.}, 1,901 GFLOPs). Meanwhile, it cannot achieve promising results for multi-level degradations. 
For efficient SR methods, light-weight CARN~\cite{ahn2018fast} achieves the best performance of 30.43db PSNR at the bicubic degradation, compared to KXNet~\cite{fu2022kxnet} and Liang \emph{et al.}~\cite{liang2022efficient}. However, similar to RRDB, CARN fails to achieves the promising results when handling the complex degradations. Our method achieves the best trade-off between PSNR and GFLOPs, compared to all methods. For example, compared to Liang \emph{et al.}~\cite{liang2022efficient}, the proposed DCS-RISR achieves a slight PSNR increase at the level-1 and level-3 degradations, while significantly reducing the 56 GFLOPs and 6.52M parameters. 
%

We further evaluate the effectiveness of our DCS-RISR on Set5, set14, B100 and Urban100 datasets with the degradation level from 1 to 3. As shown in Tab.~\ref{tab_set5}, we found that the proposed DCS-RISR achieves the best performance at multiple degradation levels and different scales on 4 benchmark datasets. Specifically, at the seriously degraded level-3, DCS-RISR almost outperforms previous SOTA SR methods on different benchmark datasets. It indicates that our method has a strong generalization ability to be effectively transfer to other domains.

\textbf{Qualitative evaluation.}
Fig~\ref{fig_6} shows the visualization results of different methods on DIV2K validation set, where the images of ID  808 and 831 are for  level-1 and level-2, respectively. Samples for bicubic and level-3 are presented in supplementaries. 
we can observe that our DCS-RISR can effectively restore rich textures and edge information, compared to other methods. In particular, our DCS-RISR reduces the effect of noise and blurring artifacts at the degradation level 1. Even under the severe degradation, DCS-RISR  can also restore image details best than other methods. For example, at the degradation level 2, we obtain the clear SR image without the blurring and noise in the window beam. More samples are presented in supplementaries.

\subsection{Ablation Study}

We conducted ablation study to evaluate the effectiveness of DCS-RISR components, including the effect of non-local regularization losses on SRResNet and dynamic structures, the number of $K$, and the element number of $\mathbf{a}$. We adopt DIV2K dataset for our ablation. The result we report is at the degradation of Level 2, unless otherwise specified.

{\bf{Effects of non-local regularization on SRResNet and dynamic structures.}} In Fig.~\ref{fig_nl}, the regression loss is removed for SRResNet and set to the same balance parameter for dynamic situation. We found that with the GAN and perceptual losses, adding the non-local regularization achieves at least 0.04db PSNR gains over the L1 reconstruction loss both on SRResNet and dynamic structures with LOC block. leveraging both L1 and non-local regularization into the GAN and perceptual losses achieves the best performance, compared to those results with NL or L1. Furthermore, we found the effectiveness of learnable LOC by reducing the about 50 GFLOPs with the relative consistent PSNR, compared to the static SRResNet.

\begin{figure}[t]
\begin{minipage}{0.48\linewidth}
\vspace{3pt}
\centerline{\includegraphics[width=\textwidth]{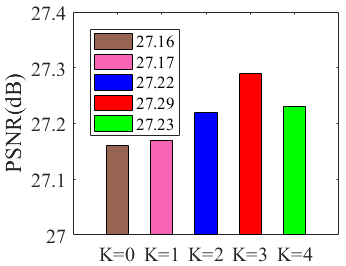}}
\centerline{\small (a) Effect of $K$ }
\end{minipage}
\begin{minipage}{0.48\linewidth}
\centerline{\includegraphics[width=\textwidth]{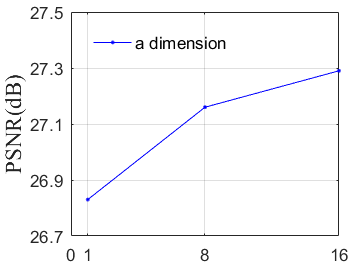}}
\centerline{\small (b) Effect of $\mathbf{a}$ dimension}
\end{minipage}
\vspace{-0.5em}
\caption{Effects of $K$ and $\mathbf{a}$ dimension. (a) Comparison results of different $K$. (b) PSNR results with different dimension of $\mathbf{a}$.}
\label{fig_a&k}
\end{figure}

{\bf{Effect of non-local patch number.}} As shown in Fig.~\ref{fig_a&k}(a), the increase of non-local path number will improve the performance, when $K\leq 3$. The performance drops when setting $K$ to 4. Note that $K=0$ indicates only calculating the query patches similarity of LR and SR images. Thus, we set $K$ to 3 for our experiments.

{\textbf{Effect of the $\mathbf{a}$ dimension.}} The dimension of $\mathbf{a}$ can be flexible to control the frequency rate of multiple LOC blocks. We further explore the effect of multiple LOC blocks sharing the same channel splitting scale. Our backbone has total 16 blocks, where we set the dimension of  $\mathbf{a}$ from a set \{1,8,16\}. As shown in Fig.~\ref{fig_a&k}(b), adding one block with a scale achieves the highest PSNR.


\section{Conclusion}
In this paper, we developed a dynamic channel splitting (DCS) method for efficient real-world image super-resolution, which simultaneously handle the problems of model redundancy and real-world complex degradation.  
We proposed a learnable octave convolution (LOC) block to dynamically allocate channel rate for different frequencies, which is controlled by the LR content via two light-weight networks. Moreover, a non-local regularization is introduced to remedy the L1-norm reconstruction loss for further performance improvement. Extensive experiments demonstrate that the proposed approach achieves the superior or comparable results over different datasets.

{\small
\bibliographystyle{ieee_fullname}
\bibliography{arxiv}
}

\section{Appendix}
In the supplementary materials, we first added more experimental results about ablation study in Sec.~\ref{sec_ablation}. Then, we provide more sample images about different degradation levels in Sec.~\ref{sec_sample}. Finally, we provide more quantitative and qualitative results compared to SOTA methods in Sec.~\ref{sec_compared}. 

\begin{table*}[t]
\centering
\caption{Quantitative results of $\times$4 on DIV2K validation set with different $\lambda_5$. GFLOPs, parameters and running times are evaluated on an average image with 256 × 256 pixels for inference using one NVIDIA 3090 GPU in all tables and figures.}
\label{tab_lam5}
\begin{tabular}{|cl|cc|cc|cc|cc|c|c|c|}
\hline
\multicolumn{2}{|c|}{\multirow{2}{*}{$\lambda_5$}} & \multicolumn{2}{c|}{Level 0}                          & \multicolumn{2}{c|}{Level 1}                          & \multicolumn{2}{c|}{Level 2}                          & \multicolumn{2}{c|}{Level 3}                          & \multicolumn{1}{c|}{Times} & \multirow{2}{*}{GFLOPs} & \multirow{2}{*}{Params(M)} \\ \cline{3-10}
\multicolumn{2}{|c|}{}                    & \multicolumn{1}{l|}{PSNR} & \multicolumn{1}{l|}{SSIM} & \multicolumn{1}{l|}{PSNR} & \multicolumn{1}{l|}{SSIM} & \multicolumn{1}{l|}{PSNR} & \multicolumn{1}{l|}{SSIM} & \multicolumn{1}{l|}{PSNR} & \multicolumn{1}{l|}{SSIM} &    (ms)                        &                           &                            \\ \hline
\multicolumn{2}{|c|}{0.25}                & 29.01                     & 0.7970                    & 27.89                     & 0.7828                    & 27.29                     & 0.7404                    & 23.95                     & 0.6156                    & 98                         & 128                       & 1.55                       \\ 
\multicolumn{2}{|c|}{0.5}                 & 28.78                     & 0.7920                    & 27.79                     & 0.7791                    & 27.20                     & 0.7361                    & 23.94                     & 0.6127                    & 95                         & 122                       & 1.55                       \\ 
\multicolumn{2}{|c|}{0.75}                & 28.58                     & 0.7863                    & 27.73                     & 0.7763                    & 27.16                     & 0.7357                    & 23.94                     & 0.6132                    & 94                         & 117                       & 1.55                       \\ 
\multicolumn{2}{|c|}{1}                   & 28.80                     & 0.8047                    & 27.64                     & 0.7853                    & 27.25                     & 0.7560                    & 23.72                     & 0.6258                    & 93                         & 108                       & 1.55                       \\ \hline
\end{tabular}
\end{table*}

\subsection{Additional Ablation Study}
\label{sec_ablation}
We further conduct ablation study to evaluate the effectiveness of DCS-RISR, including the sensitivity of sparsity hyper-parameter $\lambda_5$, the effect on different backbones and comparison with the LOC layer.

\textbf{Effect of the sparsity parameter $\lambda_5$.} $\lambda_5$ can control the rate between the high-frequency and low-frequency featue of LOC block. As shwon in Tab.~\ref{tab_lam5}, 
larger $\lambda_5$ forces the network $A$ to learn a smaller value in $\mathbf{a}$, such that reducing the usage of high-frequency features to accelerate the computation. For example, compared to the setting of $\lambda_5$ by 0.25,
setting $\lambda_5$ to 1 reduces the inference time of 5ms and 20 GFLOPs for inference. However, the PSNR significantly decreases from level-0 to level-3, \emph{i.e.}, 0.21dB, 0.26dB, 0.04dB and 0.23dB PSNR compared to those of $\lambda_5=0.25$. Consider the best trade-off between computation overhead and PSNR, we set $\lambda_5$ to 0.25 in our experiments.
%

\textbf{Effects of different backbones.}
The proposed learnable octave convolution (LOC) block can flexibly adapt to off-the-shelf SR models by replacing the residual block in the trunk part. We further verified the effect by leveraging the LOC block into different backbones, including SRResNet and EDSR. Tab.~\ref{tab_opt} shows the different structures of SRResNet and EDSR. We adopt lightweight EDSR * as backbone to compare with SRResNet for fairness.
As shown in Tab.~\ref{tab_backbone}, EDSR* with our LOC block achieves better performance. For example, EDSR* is 0.08dB higher on PSNR than SRResNet on level 3. Obviously, compared to other SOTA methods, EDSR with our LOC block achieves the best performance. 
We select SRResNet as our backbone, due to the fair comparison with Liang \emph{et al.} [30]. 

\textbf{LOC block \emph{vs.} LOC layer.} 
LOC layer is designed by directly applying the OctConv into the backbone without the merging between the outputs of high- and low-frequency features in the second convolution, \emph{i.e.}, removing the Eq.~[5]. Although LOC layer is more fine-grained than the LOC block, LOC layer achieves the worse performance compared to our LOC block. %
As shown in table~\ref{tab_stru}, LOC block achieves 15ms faster inference time and higher PSNR in all levels, comapred to that using the LOC layer.

\begin{figure*}[t]
\centering
\begin{minipage}{0.18\linewidth}
\centerline{\includegraphics[width=\textwidth]{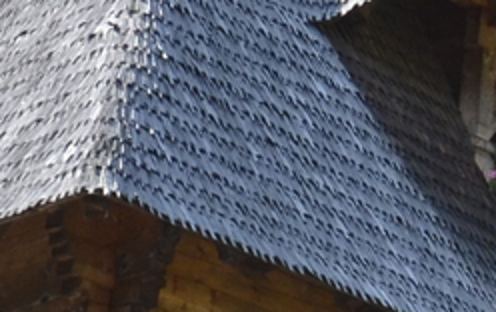}}
\vspace{3pt}
\centerline{\includegraphics[width=\textwidth]{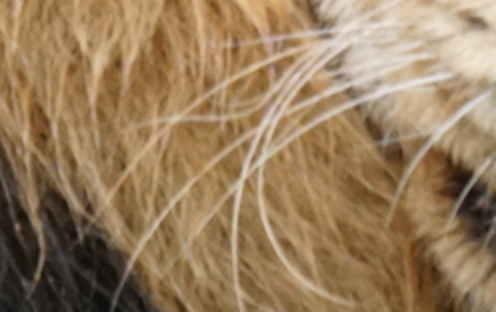}}
\vspace{3pt}
\centerline{\includegraphics[width=\textwidth]{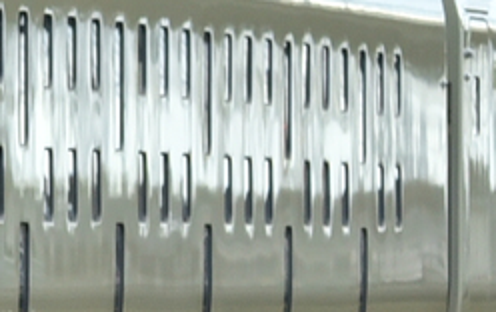}}
\vspace{3pt}
\centerline{\includegraphics[width=\textwidth]{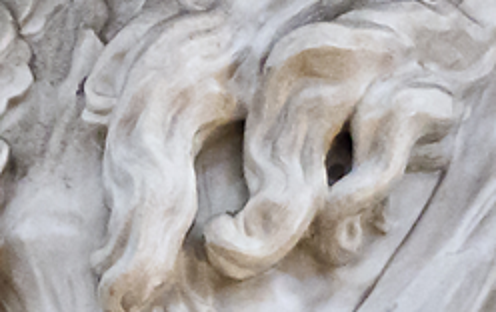}}
\vspace{3pt}
\centerline{\includegraphics[width=\textwidth]{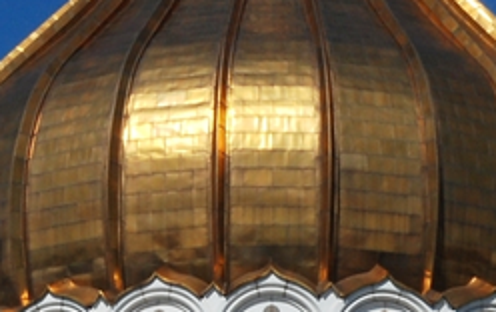}}
\vspace{3pt}
\centerline{\includegraphics[width=\textwidth]{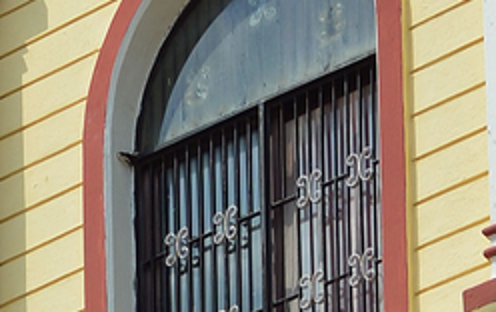}}
\vspace{3pt}
\centerline{\small (a) HR}
\end{minipage}
\begin{minipage}{0.18\linewidth}
\centerline{\includegraphics[width=\textwidth]{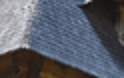}}
\vspace{3pt}
\centerline{\includegraphics[width=\textwidth]{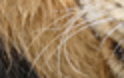}}
\vspace{3pt}
\centerline{\includegraphics[width=\textwidth]{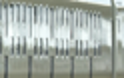}}
\vspace{3pt}
\centerline{\includegraphics[width=\textwidth]{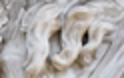}}
\vspace{3pt}
\centerline{\includegraphics[width=\textwidth]{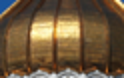}}
\vspace{3pt}
\centerline{\includegraphics[width=\textwidth]{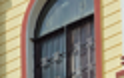}}
\vspace{3pt}
\centerline{\small (b) Bicubic}
\end{minipage}
\begin{minipage}{0.18\linewidth}
\centerline{\includegraphics[width=\textwidth]{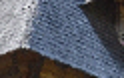}}
\vspace{3pt}
\centerline{\includegraphics[width=\textwidth]{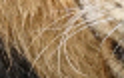}}
\vspace{3pt}
\centerline{\includegraphics[width=\textwidth]{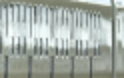}}
\vspace{3pt}
\centerline{\includegraphics[width=\textwidth]{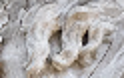}}
\vspace{3pt}
\centerline{\includegraphics[width=\textwidth]{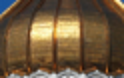}}
\vspace{3pt}
\centerline{\includegraphics[width=\textwidth]{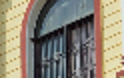}}
\vspace{3pt}
\centerline{\small (c) Level 1}
\end{minipage}
\begin{minipage}{0.18\linewidth}
\centerline{\includegraphics[width=\textwidth]{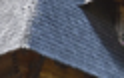}}
\vspace{3pt}
\centerline{\includegraphics[width=\textwidth]{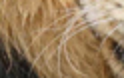}}
\vspace{3pt}
\centerline{\includegraphics[width=\textwidth]{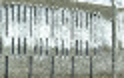}}
\vspace{3pt}
\centerline{\includegraphics[width=\textwidth]{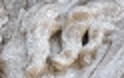}}
\vspace{3pt}
\centerline{\includegraphics[width=\textwidth]{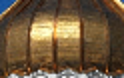}}
\vspace{3pt}
\centerline{\includegraphics[width=\textwidth]{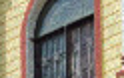}}
\vspace{3pt}
\centerline{\small (d) Level 2}
\end{minipage}
\begin{minipage}{0.18\linewidth}
\centerline{\includegraphics[width=\textwidth]{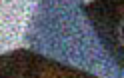}}
\vspace{3pt}
\centerline{\includegraphics[width=\textwidth]{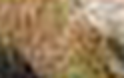}}
\vspace{3pt}
\centerline{\includegraphics[width=\textwidth]{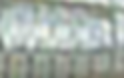}}
\vspace{3pt}
\centerline{\includegraphics[width=\textwidth]{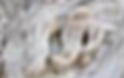}}
\vspace{3pt}
\centerline{\includegraphics[width=\textwidth]{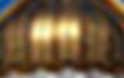}}
\vspace{3pt}
\centerline{\includegraphics[width=\textwidth]{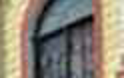}}
\vspace{3pt}
\centerline{\small (e) Level 3}
\end{minipage}
\vspace{-.5em}
\caption{More sample images from DIV2K validation set with different degradation levels.}
\label{fig_sample_supp}
\end{figure*}

\begin{table}[t]
\centering
\small
\caption{The architecture of the SRResNet and EDSR. EDSR* denote the lightweight EDSR.}
\label{tab_opt}
\begin{tabular}{l|ccc}
Options            & \multicolumn{1}{l}{SRResNet} & \multicolumn{1}{l}{EDSR*} & \multicolumn{1}{l}{EDSR} \\ \hline
\# Residual blocks & 16                           & 16                        & 32                       \\
\# Filters         & 64                           & 64                        & 256                      \\
\# Parameters      & 1.52M                        & 1.52M                     & 43M                     
\end{tabular}
\end{table}

\begin{table*}[t]
\centering
\small
\caption{Quantitative results of $\times$4 SR with different backbones on DIV2K validation set. }
\label{tab_backbone}
\begin{tabular}{|cl|ll|ll|ll|ll|c|c|c|}
\hline
\multicolumn{2}{|c|}{\multirow{2}{*}{Backbone}} & \multicolumn{2}{c|}{Level 0}                            & \multicolumn{2}{c|}{Level 1}                            & \multicolumn{2}{c|}{Level 2}                            & \multicolumn{2}{c|}{Level 3}                            & \multicolumn{1}{c|}{Times} & \multirow{2}{*}{GFLOPs} & \multirow{2}{*}{Params(M)} \\ \cline{3-10}
\multicolumn{2}{|c|}{}                          & \multicolumn{1}{l|}{PSNR} & SSIM                        & \multicolumn{1}{l|}{PSNR} & SSIM                        & \multicolumn{1}{l|}{PSNR} & SSIM                        & \multicolumn{1}{l|}{PSNR} & SSIM                        &     (ms)                       &                         &                            \\ \hline
\multicolumn{2}{|c|}{SRResNet}                  & \multicolumn{1}{c}{29.01} & \multicolumn{1}{c|}{0.7970} & \multicolumn{1}{c}{27.89} & \multicolumn{1}{c|}{0.7828} & \multicolumn{1}{c}{27.29} & \multicolumn{1}{c|}{0.7404} & \multicolumn{1}{c}{23.95} & \multicolumn{1}{c|}{0.6156} & 98                         & 128                     & 1.55                       \\
\multicolumn{2}{|c|}{EDSR*}                      &28.69 &	0.7877 &	\multicolumn{1}{c}{27.93} &	0.7891 	&\multicolumn{1}{c}{27.34} &	0.7449 &	\multicolumn{1}{c}{24.03} &	0.6251 			
                       & \multicolumn{1}{c|}{82}      & \multicolumn{1}{c|}{101}   & \multicolumn{1}{c|}{1.52}      \\ \hline
\end{tabular}

\end{table*}

\begin{table*}[t]
\centering
\small
\caption{comparison between LOC block and LOC layer for $\times$4 SR on DIV2K validation set.}
\label{tab_stru}
\begin{tabular}{|cl|ll|ll|ll|ll|c|c|c|}
\hline
\multicolumn{2}{|c|}{\multirow{2}{*}{Method}} & \multicolumn{2}{c|}{Level 0}                            & \multicolumn{2}{c|}{Level 1}                            & \multicolumn{2}{c|}{Level 2}                            & \multicolumn{2}{c|}{Level 3}                            & \multicolumn{1}{c|}{Times} & \multirow{2}{*}{GFLOPs} & \multirow{2}{*}{Params(M)} \\ \cline{3-10}
\multicolumn{2}{|c|}{}                          & \multicolumn{1}{l|}{PSNR} & SSIM                        & \multicolumn{1}{l|}{PSNR} & SSIM                        & \multicolumn{1}{l|}{PSNR} & SSIM                        & \multicolumn{1}{l|}{PSNR} & SSIM                        &     (ms)                       &                         &                            \\ \hline
\multicolumn{2}{|c|}{LOC block}                  & \multicolumn{1}{c}{29.01} & \multicolumn{1}{c|}{0.7970} & \multicolumn{1}{c}{27.89} & \multicolumn{1}{c|}{0.7828} & \multicolumn{1}{c}{27.29} & \multicolumn{1}{c|}{0.7404} & \multicolumn{1}{c}{23.95} & \multicolumn{1}{c|}{0.6156} & 98                         & 128                     & 1.55                       \\
\multicolumn{2}{|c|}{LOC layer}                      &27.49 &	0.7560 	&27.27 &	0.7582& 	26.73& 	0.7279 &	23.94 &	0.6210& 		
\multicolumn{1}{c|}{113}      & \multicolumn{1}{c|}{144}   & \multicolumn{1}{c|}{1.55}      \\ \hline
\end{tabular}
\end{table*}

\subsection{More Sample Images from Degradation Levels}
\label{sec_sample}
As shown in Fig.~\ref{fig_sample_supp}, We provide more sample images from different degradation levels. The images from the degradation Level 1 to 3 cover a variety of degradation from simple patterns to complex ones. Therefore, the restoration of images from higher degradation layers is more difficult.

\subsection{More Results Compared to SOTA Methods}
\label{sec_compared}
To further verify the effectiveness of DCS-RISR, we provide more quantitative and quantitative comparisons with previous SOTA methods.

\textbf{More quantitative comparisons.}
Tab.~\ref{tab_PSNR} and tab.~\ref{tab_SSIM} show the PSNR and SSIM results on four benchmark datasets with the degradation level from 1 to 3, respectively. In the main paper, we have evaluated the effectiveness for $\times$2 and $\times$4 SR performance. $\times$3 results are summarized in Tab.~\ref{tab_PSNR}. 
We can see that DCS-RISR achieves the highest PSNR across three levels on Set14, compared to other SOTA methods. In other datasets, we achieves the comparable results.
For SSIM in Tab.~\ref{tab_SSIM}, DCS-RISR achieves the most best ranking performance for $\times$2 and $\times$4 SR on four benchmark datasets across level 1 to 3. For example, on set14 and Urban100, DCS-RISR achieves about 0.068 and 0.056 SSIM gains over the second-best ranking methods on $\times$2 SR. 
These results indicate the effectiveness of our DCS-RISR by dynamic channel splitting scheme. 

\begin{figure*}[t]
\centering
\begin{minipage}{0.3\linewidth}
\centerline{\includegraphics[width=\textwidth]{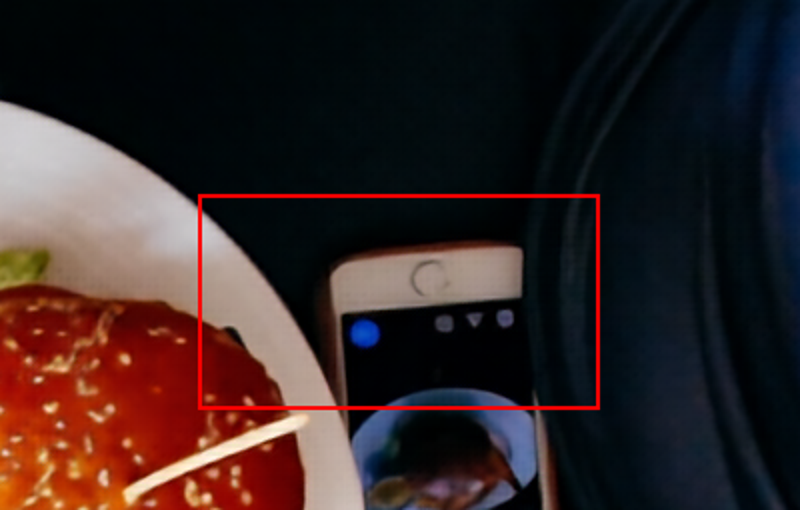}}
\centerline{\small (1) Bicubic 0848.png}
\vspace{2pt}
\centerline{\includegraphics[width=\textwidth]{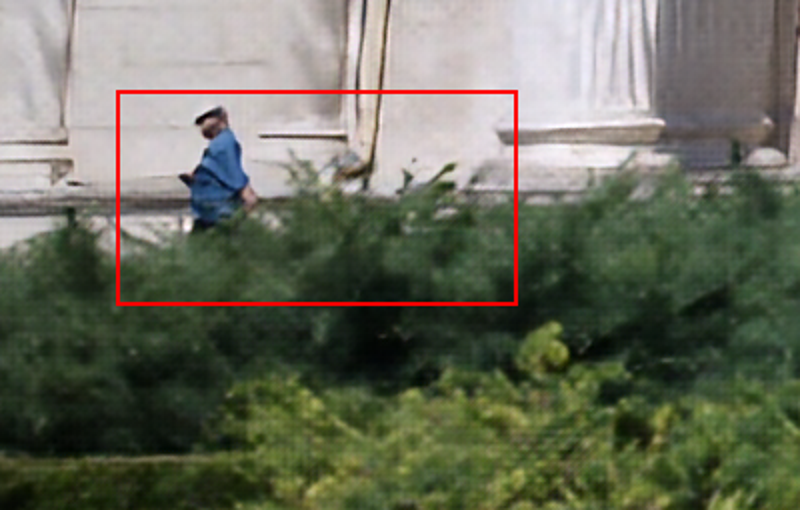}}
\centerline{\small (2) Level 1 0812.png}
\vspace{2pt}
\centerline{\includegraphics[width=\textwidth]{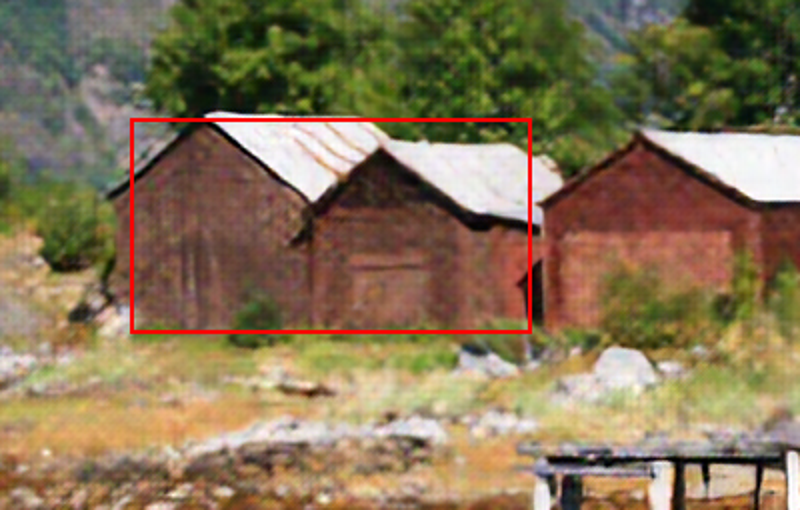}}
\centerline{\small (3) Level 2 0808.png}
\vspace{2pt}
\centerline{\includegraphics[width=\textwidth]{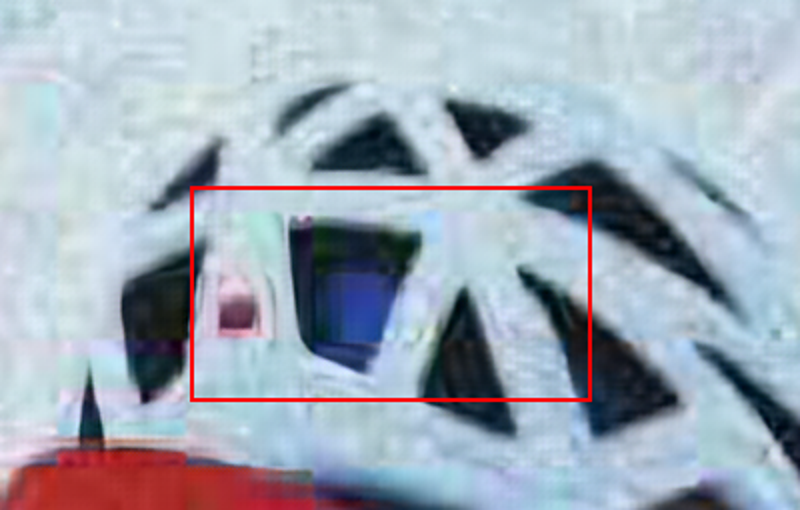}}
\centerline{\small (4) Level 3 0844.png}
\vspace{2pt}
\end{minipage}
\begin{minipage}{0.149\linewidth}
\centerline{\includegraphics[width=\textwidth]{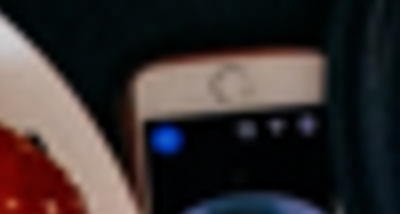}}
\centerline{\small (a) Bicubic}
\vspace{2pt}
\centerline{\includegraphics[width=\textwidth]{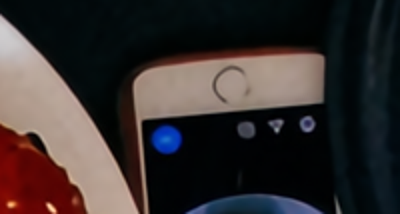}}
\centerline{\small (e) DAN}
\vspace{2pt}
\centerline{\includegraphics[width=\textwidth]{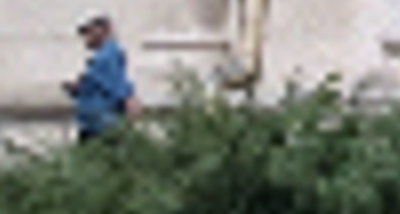}}
\centerline{\small (a) Bicubic}
\vspace{2pt}
\centerline{\includegraphics[width=\textwidth]{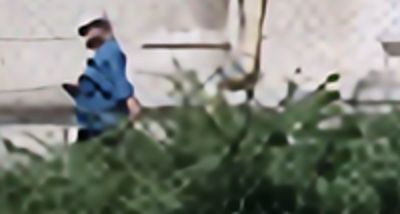}}
\centerline{\small (e) DAN}
\vspace{2pt}
\centerline{\includegraphics[width=\textwidth]{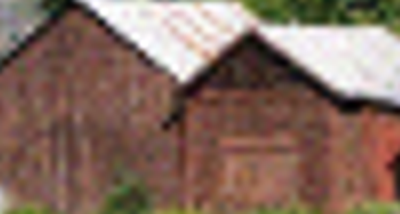}}
\centerline{\small (a) Bicubic}
\vspace{2pt}
\centerline{\includegraphics[width=\textwidth]{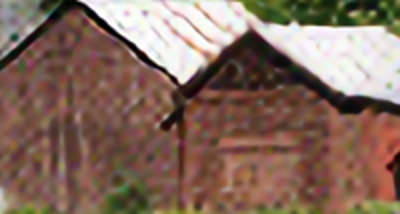}}
\centerline{\small (e) DAN}
\vspace{2pt}
\centerline{\includegraphics[width=\textwidth]{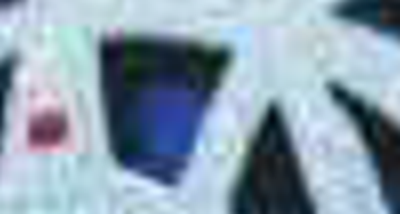}}
\centerline{\small (a) Bicubic}
\vspace{2pt}
\centerline{\includegraphics[width=\textwidth]{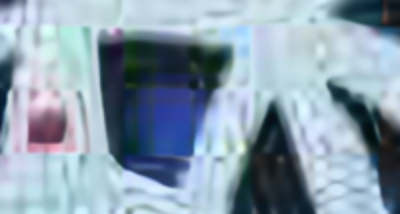}}
\centerline{\small (e) DAN}
\vspace{2pt}
\end{minipage}
\begin{minipage}{0.149\linewidth}
\centerline{\includegraphics[width=\textwidth]{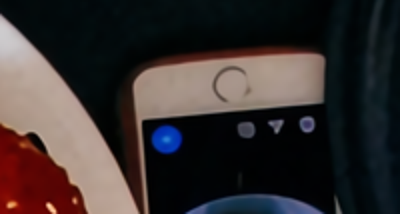}}
\centerline{\small (b) RRDB}
\vspace{2pt}
\centerline{\includegraphics[width=\textwidth]{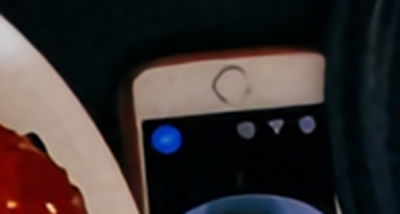}}
\centerline{\small (f) DASR}
\vspace{2pt}
\centerline{\includegraphics[width=\textwidth]{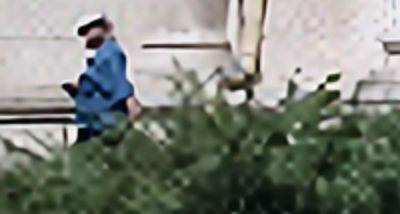}}
\centerline{\small (b) RRDB}
\vspace{2pt}
\centerline{\includegraphics[width=\textwidth]{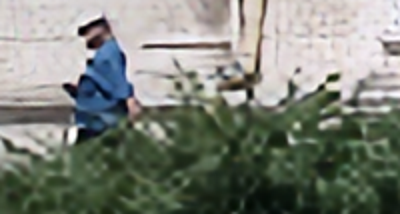}}
\centerline{\small (f) DASR}
\vspace{2pt}
\centerline{\includegraphics[width=\textwidth]{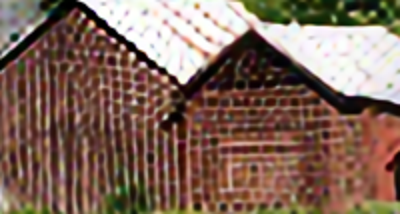}}
\centerline{\small (b) RRDB}
\vspace{2pt}
\centerline{\includegraphics[width=\textwidth]{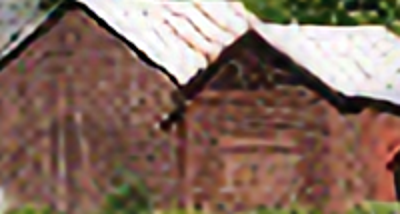}}
\centerline{\small (f) DASR}
\vspace{2pt}
\centerline{\includegraphics[width=\textwidth]{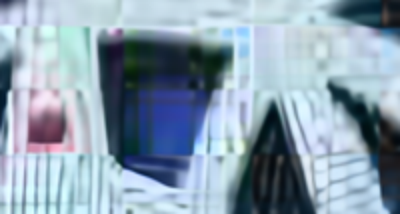}}
\centerline{\small (b) RRDB}
\vspace{2pt}
\centerline{\includegraphics[width=\textwidth]{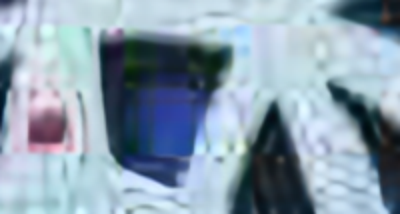}}
\centerline{\small (f) DASR}
\vspace{2pt}
\end{minipage}
\begin{minipage}{0.149\linewidth}
\centerline{\includegraphics[width=\textwidth]{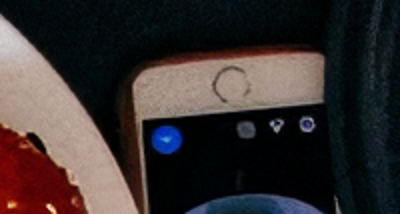}}
\centerline{\small (c) ESRGAN}
\vspace{2pt}
\centerline{\includegraphics[width=\textwidth]{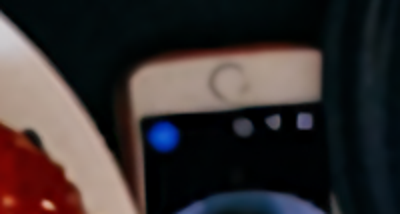}}
\centerline{\small (g) KXNet}
\vspace{2pt}
\centerline{\includegraphics[width=\textwidth]{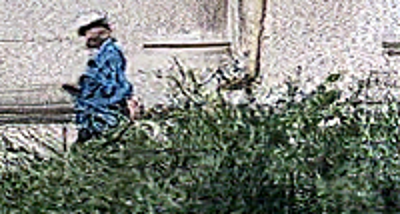}}
\centerline{\small (c) ESRGAN}
\vspace{2pt}
\centerline{\includegraphics[width=\textwidth]{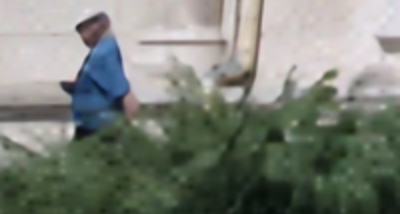}}
\centerline{\small (g) KXNet}
\vspace{2pt}
\centerline{\includegraphics[width=\textwidth]{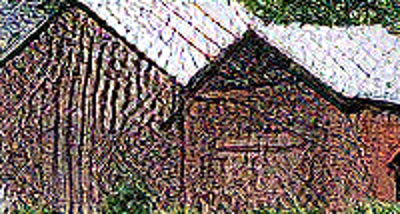}}
\centerline{\small (c) ESRGAN}
\vspace{2pt}
\centerline{\includegraphics[width=\textwidth]{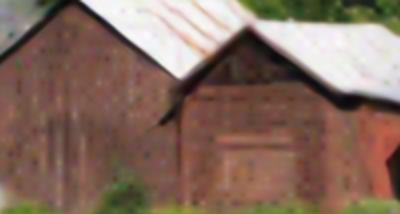}}
\centerline{\small (g) KXNet}
\vspace{2pt}
\centerline{\includegraphics[width=\textwidth]{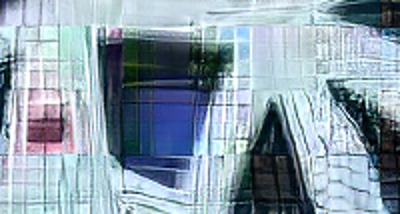}}
\centerline{\small (c) ESRGAN}
\vspace{2pt}
\centerline{\includegraphics[width=\textwidth]{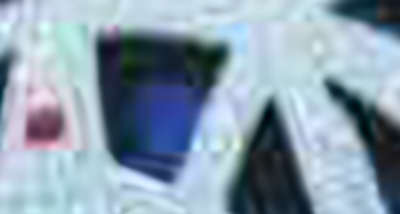}}
\centerline{\small (g) KXNet}
\vspace{2pt}
\end{minipage}
\begin{minipage}{0.149\linewidth}
\centerline{\includegraphics[width=\textwidth]{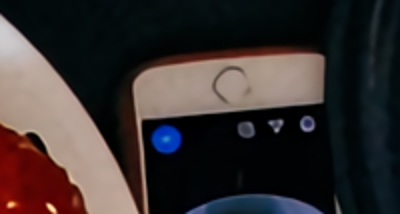}}
\centerline{\small (d) IKC}
\vspace{2pt}
\centerline{\includegraphics[width=\textwidth]{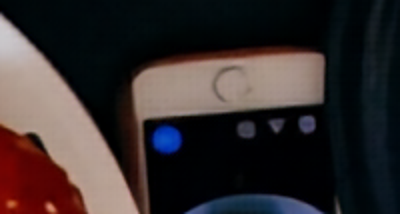}}
\centerline{\small (h) DCS-RISR}
\vspace{2pt}
\centerline{\includegraphics[width=\textwidth]{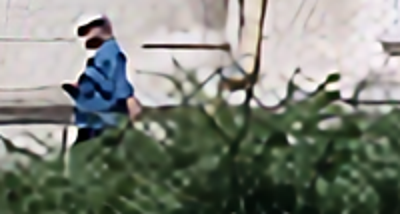}}
\centerline{\small (d) IKC}
\vspace{2pt}
\centerline{\includegraphics[width=\textwidth]{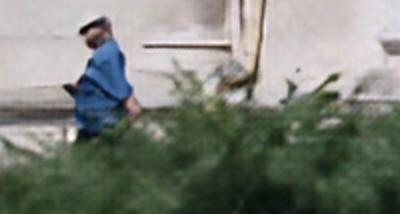}}
\centerline{\small (h) DCS-RISR}
\vspace{2pt}
\centerline{\includegraphics[width=\textwidth]{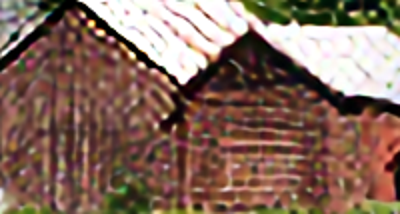}}
\centerline{\small (d) IKC}
\vspace{2pt}
\centerline{\includegraphics[width=\textwidth]{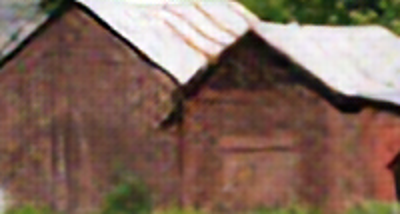}}
\centerline{\small (h) DCS-RISR}
\vspace{2pt}
\centerline{\includegraphics[width=\textwidth]{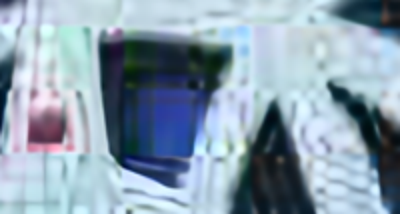}}
\centerline{\small (d) IKC}
\vspace{2pt}
\centerline{\includegraphics[width=\textwidth]{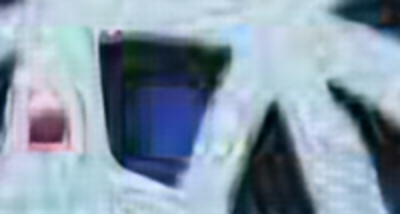}}
\centerline{\small (h) DCS-RISR}
\vspace{2pt}
\end{minipage}
\vspace{-0.5em}
\caption{Qualitative Comparison at different degradation levels of $\times$4 on DIV2K validation set. Zoom in for best views.}
\label{fig_vis}
\end{figure*}

\textbf{More qualitative comparisons.} 
In Fig.~\ref{fig_vis}, we have provided more SR visualization results by degradation from bicubic, Level 1, Level 2 and Level 3 on DIV2K validation set. For example, ESRGAN amplifies the noise around the house and KXNet blurs the image, while IKC, DAN and DASR produce detail artifacts on the degradation level 1 (Fig.~\ref{fig_vis}(2)) and level 2 (Fig.~\ref{fig_vis}(3)). On the severe degradation level 3, the previous SOTA methods cannot effectively remove noise and loss the details. In contrast, the proposed DCS-RISR is able to enhance the LR images and achieves the best visualization results on all degradation levels.

\begin{table*}[t]
\centering
\small
\caption{ PSNR results on Set5, Set14, B100 and Urban100 from the degradation level S1-S3.
L1, L2 and L3 mean Level-1, Level-2 and Level-3,respectively. \textbf{Bold} and \underline{underline} number are the best and ranking-second best performance in all tables.}
\label{tab_PSNR}
\begin{tabular}{|cl|c|lll|lll|lll|lll|}
\hline
\multicolumn{2}{|c|}{\multirow{2}{*}{Method}} & \multirow{2}{*}{scale} & \multicolumn{3}{c|}{Set5}                                                                  & \multicolumn{3}{c|}{Set14}                                                                 & \multicolumn{3}{c|}{B100}                                                                  & \multicolumn{3}{c|}{Urban100}                                                              \\ \cline{4-15} 
\multicolumn{2}{|c|}{}                        &                        & \multicolumn{1}{c|}{L 1} & \multicolumn{1}{c|}{L 2} & \multicolumn{1}{c|}{L 3} & \multicolumn{1}{c|}{L 1} & \multicolumn{1}{c|}{L 2} & \multicolumn{1}{c|}{L 3} & \multicolumn{1}{c|}{L 1} & \multicolumn{1}{c|}{L 2} & \multicolumn{1}{c|}{L 3} & \multicolumn{1}{c|}{L 1} & \multicolumn{1}{c|}{L 2} & \multicolumn{1}{c|}{L 3} \\ \hline
\multicolumn{2}{|c|}{SRResNet [25]}                 & \multirow{8}{*}{×3}    & 28.26                        & 27.38                        & 22.80                        & 25.46                        & 26.10                        & \underline{22.80}                        & 25.67                        & 25.56                        & 23.00                        & 22.81                        & 23.32                        & 20.63                        \\ \cline{1-2}
\multicolumn{2}{|c|}{RCAN [56]}                    &                        & 28.25                        & 27.32                        & 22.80                        & 25.26                        & 26.14                        & 22.78                        & 25.67                        & 25.55                        & 22.98                        & 22.51                        & 23.27                        & 20.62                        \\ \cline{1-2}
\multicolumn{2}{|c|}{IKC [19]}                     &                        & \textbf{29.79}               & \textbf{28.64}               & 23.00                        & \underline{26.71}                        & \underline{26.52}                        & 22.56                        & \textbf{26.63}               & 25.84                        & 22.53                        & \textbf{24.07}               & \underline{24.02}                        & 20.43                        \\ \cline{1-2}
\multicolumn{2}{|c|}{DAN [22]}                     &                        & 29.48                        & \underline{28.43}                        & 22.97                        & 26.51                        & 26.51                        & 22.77                        & 26.36                        & \underline{25.92}                        & 22.88                        & 23.74                        & 23.90                        & \textbf{20.68}                        \\ \cline{1-2}
\multicolumn{2}{|c|}{DASR [44]}                    &                        & 28.46                        & 27.70                        & 22.23                        & 25.68                        & 26.31                        & 22.72                        & 25.91                        & 25.71                        & 22.67                        & 23.05                        & 23.49                        & 20.61                        \\ \cline{1-2}
\multicolumn{2}{|c|}{CARN [2]}                    &                        & 28.33                        & 27.43                        & 22.81                        & 25.60                        & 26.14                        & \underline{22.80}                        & 25.75                        & 25.58                        & \underline{23.01}                        & 22.95                        & 23.37                        & 20.64                        \\ \cline{1-2}
\multicolumn{2}{|c|}{KXNet [16]}                   &                        & 24.28                        & 24.62                        & \textbf{23.58}                        & 24.46                        & 24.81                        & 22.62                        & 24.84                        & 25.10                        & 22.61                        & 22.00                        & 22.40                        & 20.49                        \\ \cline{1-2}
\multicolumn{2}{|c|}{DCS-RISR}                &                        & \underline{29.49}                        & 28.25                        & \underline{23.09}               & \textbf{26.87}               & \textbf{26.57}               & \textbf{22.89}               & \underline{26.27}                        & \textbf{25.93}               & \textbf{23.14}               & \underline{23.82}                        & \textbf{24.11}               & \underline{20.65}               \\ \hline
\end{tabular}
\end{table*}

\begin{table*}[t]
\centering
\footnotesize
\caption{SSIM results on Set5, Set14, B100 and Urban100 from the degradation level S1-S3.}
\label{tab_SSIM}
\begin{tabular}{|cl|c|lll|lll|lll|lll|}
\hline
\multicolumn{2}{|c|}{\multirow{2}{*}{Method}} & \multirow{2}{*}{scale} & \multicolumn{3}{c|}{Set5}                                                      & \multicolumn{3}{c|}{Set14}                                                     & \multicolumn{3}{c|}{B100}                                                      & \multicolumn{3}{c|}{Urban100}                                                  \\ \cline{4-15} 
\multicolumn{2}{|c|}{}                        &                        & \multicolumn{1}{c|}{L 1} & \multicolumn{1}{c|}{L 2} & \multicolumn{1}{c|}{L 3} & \multicolumn{1}{c|}{L 1} & \multicolumn{1}{c|}{L 2} & \multicolumn{1}{c|}{L 3} & \multicolumn{1}{c|}{L 1} & \multicolumn{1}{c|}{L 2} & \multicolumn{1}{c|}{L 3} & \multicolumn{1}{c|}{L 1} & \multicolumn{1}{c|}{L 2} & \multicolumn{1}{c|}{L 3} \\ \hline
\multicolumn{2}{|c|}{SRResNet [25]}                & \multirow{8}{*}{×2}    & 0.9175                   & 0.8227                   & 0.5699                   & 0.8617                   & 0.7432                   & 0.4761                   & 0.8246                   & 0.6672                   & 0.4923                   & 0.8408                   & 0.7026                   & 0.4859                   \\ \cline{1-2}
\multicolumn{2}{|c|}{RCAN [56]}                    &                        & {\underline{0.9187}}             & 0.8230                   & 0.5683                   & 0.8613                   & 0.7426                   & 0.4750                   & 0.8253                   & 0.6660                   & 0.4924                   & 0.8403                   & 0.7034                   & 0.4858                   \\ \cline{1-2}
\multicolumn{2}{|c|}{IKC [19]}                     &                        & 0.9158                   & 0.8294                   & 0.5556                   & 0.8640                   & 0.7505                   & 0.4591                   & \textbf{0.8326}          & 0.6866                   & 0.4806                   & {\underline{0.8512}}             & 0.7169                   & 0.4770                   \\ \cline{1-2}
\multicolumn{2}{|c|}{DAN [22]}                     &                        & 0.9186                   & 0.8285                   & 0.5527                   & {\underline{0.8653}}             & 0.7473                   & 0.4693                   & {\underline{0.8310}}             & 0.6717                   & 0.4774                   & 0.8489                   & 0.7098                   & 0.4774                   \\ \cline{1-2}
\multicolumn{2}{|c|}{DASR [44]}                    &                        & 0.8495                   & 0.8239                   & 0.5674                   & 0.8643                   & 0.7480                   & 0.4648                   & 0.8290                   & 0.6723                   & 0.4782                   & 0.8470                   & 0.7073                   & 0.4751                   \\ \cline{1-2}
\multicolumn{2}{|c|}{CARN [2]}                    &                        & 0.9172                   & 0.8229                   & 0.5700                   & 0.8618                   & 0.7435                   & {\underline{0.4767}}             & 0.8240                   & 0.6674                   & {\underline{0.4928}}             & 0.8405                   & 0.7020                   & 0.4862                   \\ \cline{1-2}
\multicolumn{2}{|c|}{KXNet [16]}                   &                        & 0.8775                   & {\underline{0.8548}}             & {\underline{0.5904}}             & 0.8239                   & {\underline{0.7530}}             & 0.4750                   & 0.7999                   & {\underline{0.7038}}             & 0.4877                   & 0.8093                   & {\underline{0.7303}}             & {\underline{0.4871}}             \\ \cline{1-2}
\multicolumn{2}{|c|}{DCS-RISR}                &                        & \textbf{0.9206}          & \textbf{0.8692}          & \textbf{0.6308}          & \textbf{0.8668}          & \textbf{0.7900}          & \textbf{0.5447}          & 0.8271                   & \textbf{0.7374}          & \textbf{0.5454}          & \textbf{0.8604}          & \textbf{0.7778}          & \textbf{0.5430}          \\ \hline
\multicolumn{2}{|c|}{SRResNet [25]}                & \multirow{8}{*}{×3}    & 0.7968                   & 0.6989                   & 0.5799                   & 0.7073                   & 0.7037                   & 0.5400                   & 0.7025                   & 0.6419                   & 0.5101                   & 0.6940                   & 0.6556                   & 0.4929                   \\ \cline{1-2}
\multicolumn{2}{|c|}{RCAN [56]}                    &                        & 0.7915                   & 0.6946                   & 0.5789                   & 0.7011                   & 0.7063                   & 0.5389                   & 0.7035                   & 0.6415                   & 0.5089                   & 0.6855                   & 0.6528                   & 0.4917                   \\ \cline{1-2}
\multicolumn{2}{|c|}{IKC [19]}                     &                        & \textbf{0.8368}          & {\underline{0.7516}}             & 0.5810                   & \textbf{0.7467}          & {\underline{0.7197}}             & 0.5254                   & \textbf{0.7266}          & \textbf{0.6627}          & 0.4784                   & \textbf{0.7329}          & \textbf{0.6844}          & 0.4738                   \\ \cline{1-2}
\multicolumn{2}{|c|}{DAN [22]}                     &                        & {\underline{0.8311}}             & 0.7360                   & 0.5839                   & 0.7381                   & 0.7216                   & 0.5377                   & {\underline{0.7231}}             & {\underline{0.6576}}             & 0.4976                   & {\underline{0.7237}}             & {\underline{0.6789}}             & 0.4909                   \\ \cline{1-2}
\multicolumn{2}{|c|}{DASR [44]}                    &                        & 0.7764                   & 0.7102                   & 0.5510                   & 0.6973                   & 0.7066                   & 0.5335                   & 0.7040                   & 0.6467                   & 0.4856                   & 0.6827                   & 0.6538                   & 0.4832                   \\ \cline{1-2}
\multicolumn{2}{|c|}{CARN [2]}                    &                        & 0.7987                   & 0.6998                   & 0.5814                   & 0.7097                   & 0.7042                   & {\underline{0.5404}}             & 0.7042                   & 0.6425                   & {\underline{0.5107}}             & 0.6963                   & 0.6559                   & {\underline{0.4934}}             \\ \cline{1-2}
\multicolumn{2}{|c|}{KXNet [16]}                   &                        & 0.7406                   & 0.7149                   & \textbf{0.6443}          & 0.6911                   & 0.6819                   & 0.5344                   & 0.6683                   & 0.6457                   & 0.4897                   & 0.6609                   & 0.6466                   & 0.4855                   \\ \cline{1-2}
\multicolumn{2}{|c|}{DCS-RISR}                &                        & 0.8267                   & \textbf{0.7839}          & {\underline{0.6087}}             & {\underline{0.7385}}             & \textbf{0.7385}          & \textbf{0.5499}          & 0.7089                   & 0.6568                   & \textbf{0.5253}          & 0.7029                   & 0.6571                   & \textbf{0.5038}          \\ \hline
\multicolumn{2}{|c|}{SRResNet [25]}                & \multirow{9}{*}{×4}    & 0.8331                   & 0.7052                   & 0.5566                   & 0.7040                   & 0.6531                   & 0.5152                   & {\underline{0.6597}}             & 0.6224                   & 0.4853                   & 0.6762                   & 0.6259                   & 0.4642                   \\ \cline{1-2}
\multicolumn{2}{|c|}{RCAN [56]}                    &                        & 0.8217                   & 0.6908                   & 0.5498                   & 0.6808                   & 0.6435                   & 0.5124                   & 0.6515                   & 0.6189                   & 0.4790                   & 0.6592                   & 0.6162                   & 0.4599                   \\ \cline{1-2}
\multicolumn{2}{|c|}{IKC [19]}                     &                        & 0.8450                   & 0.7260                   & 0.5343                   & 0.7137                   & 0.6709                   & 0.5054                   & 0.6676                   & 0.6331                   & 0.4723                   & 0.6843                   & 0.6373                   & 0.4544                   \\ \cline{1-2}
\multicolumn{2}{|c|}{DAN [22]}                     &                        & {\underline{0.8488}}             & {\underline{0.7444}}             & 0.5692                   & {\underline{0.7276}}             & {\underline{0.6781}}             & {\underline{0.5165}}             & 0.6800                   & {\underline{0.6424}}             & 0.4768                   & {\underline{0.7022}}             & {\underline{0.6544}}             & 0.4616                   \\ \cline{1-2}
\multicolumn{2}{|c|}{DASR [44]}                    &                        & 0.8370                   & 0.7179                   & {\underline{0.5699}}             & 0.7165                   & 0.6650                   & 0.5094                   & 0.6683                   & 0.6308                   & 0.4673                   & 0.6926                   & 0.6421                   & 0.4529                   \\ \cline{1-2}
\multicolumn{2}{|c|}{CARN [2]}                    &                        & 0.8323                   & 0.7023                   & 0.5599                   & 0.7067                   & 0.6547                   & 0.5164                   & \textbf{0.6615}          & 0.6227                   & {\underline{0.4865}}             & 0.6798                   & 0.6283                   & {\underline{0.4653}}             \\ \cline{1-2}
\multicolumn{2}{|c|}{ARM [7]}                     &                        & 0.7913                   & 0.6489                   & 0.4490                   & 0.6690                   & 0.5992                   & 0.4364                   & 0.6482                   & 0.5976                   & 0.4435                   & 0.6448                   & 0.5867                   & 0.4006                   \\ \cline{1-2}
\multicolumn{2}{|c|}{KXNet [16]}                   &                        & 0.7362                   & 0.7131                   & 0.5622                   & 0.6216                   & 0.6213                   & 0.5094                   & 0.5960                   & 0.5914                   & 0.4803                   & 0.5739                   & 0.5787                   & 0.4594                   \\ \cline{1-2}
\multicolumn{2}{|c|}{DCS-RISR}                &                        & \textbf{0.8496}          & \textbf{0.7939}          & \textbf{0.5731}          & \textbf{0.7294}          & \textbf{0.6927}          & \textbf{0.5225}          & 0.6526                   & \textbf{0.6526}          & \textbf{0.4956}          & \textbf{0.7155}          & \textbf{0.6851}          & \textbf{0.4797}          \\ \hline
\end{tabular}
\end{table*}

\end{document}